\pdfoutput=1
\documentclass{article}
\usepackage[accepted]{icml2019}
\usepackage{subfiles}

\usepackage[utf8]{inputenc}
\usepackage[T1]{fontenc}
\usepackage[
        activate={true,nocompatibility},
        tracking=true
        kerning=true,
        spacing=true
]{microtype}

\usepackage{hanging}
\usepackage{xcolor}

\usepackage{amsthm}
\usepackage{amssymb}
\usepackage{amsmath}
\usepackage{amsfonts}
\usepackage{etoolbox}
\usepackage{mathtools}
\usepackage{nicefrac}
\usepackage{siunitx}
\usepackage{stackengine}
\usepackage{tabstackengine}

\AtBeginEnvironment{pmatrix*}{\setlength{\arraycolsep}{1pt}}

\usepackage{algorithm}
\usepackage[noend]{algpseudocode}
\usepackage{listings}

\algrenewcommand\alglinenumber[1]{\tiny #1}

\lstdefinestyle{pythonStyle}{
	language=Python,
	basicstyle=\ttfamily\scriptsize,
	frame=lines,
	tabsize=4,
	numbers=left,
	numberstyle=\ttfamily\scriptsize}

\usepackage{booktabs}
\usepackage{multirow}
\usepackage{graphicx}
\usepackage[font=scriptsize]{caption}
\usepackage[font=scriptsize]{subfig}
\usepackage{natbib}
\usepackage{hyperref}
\usepackage{url}

\usepackage{xr}
\newcommand{\smallmath}[1]{\text{\tiny{$#1$}}}

\renewcommand{\leq}{\leqslant}
\renewcommand{\geq}{\geqslant}

\newcommand{\reals}{\mathbb{R}}

\newcommand{\E}{\operatorname{E}}

\newcommand{\diag}{\operatorname{diag}}
\newcommand{\vspan}{\operatorname{span}}
\newcommand{\rank}{\operatorname{rank}}

\DeclarePairedDelimiterX{\inner}[2]{\langle}{\rangle}{#1, #2}
\DeclarePairedDelimiterX{\norm}[1]{\|}{\|}{#1}

\newcolumntype{N}{S[round-mode=places,round-precision=2]}

\newcommand{\jg}{J_G}

\newcommand{\enc}{\operatorname{Enc}}

\newcommand{\pdata}{p_\text{data}}

\icmltitlerunning{A Spectral Regularizer for Unsupervised Disentanglement}
\begin{document}

\twocolumn[
\icmltitle{A Spectral Regularizer for Unsupervised Disentanglement}

\begin{icmlauthorlist}
\icmlauthor{Aditya Ramesh}{openai,formerNYU}
\icmlauthor{Youngduck Choi}{yale}
\icmlauthor{Yann LeCun}{nyu}
\end{icmlauthorlist}

\icmlaffiliation{openai}{OpenAI}
\icmlaffiliation{formerNYU}{Continuation of work done at New York University}
\icmlaffiliation{yale}{Department of Computer Science, Yale University}
\icmlaffiliation{nyu}{Department of Computer Science, New York University}

\icmlcorrespondingauthor{Aditya Ramesh}{aramesh@openai.com}
\icmlkeywords{Machine Learning, ICML, Disentangled Representations, Generative Modeling}

\vskip 0.3in
]

\printAffiliationsAndNotice{}

\begin{abstract}
A generative model with a disentangled representation allows for control over independent aspects of
the output. Learning disentangled representations has been a recent topic of great interest, but
it remains poorly understood. We show that even for~GANs that do not possess disentangled
representations, one can find curved trajectories in latent space over which local disentanglement
occurs. These trajectories are found by iteratively following the leading right-singular vectors of
the Jacobian of the generator with respect to its input. Based on this insight, we describe an
efficient regularizer that aligns these vectors with the coordinate axes, and show that it can be
used to induce disentangled representations in~GANs, in a completely unsupervised manner.
\end{abstract}

\section{Introduction}

Rapid progress has been made in the development of generative models capable of producing realistic
samples from distributions over complex, high-resolution natural images~(\citet{ProgressiveGAN},
\citet{BigGAN}). Despite the success of these models, it is unclear how one can achieve control over
the data-generation process when labeled instances are not available. Perturbing an
individual component of a latent variable typically results in an unpredictable change to the
output. When the model has a disentangled representation, these changes become interpretable, and
each component of the latent variable affects a distinct attribute. So far, the problem of learning
disentangled reprersentations remains poorly understood. Most approaches for doing this have focused
on~VAEs~(\citet{AEVB}, \citet{StochasticBackprop}), which produce blurry samples in practice. The few
approaches that have been developed for GANs, such as InfoGAN~\citep{InfoGAN}, have had comparably
limited success.

\begin{figure}[t]
\newcommand{\image}[1]{\includegraphics[width=\linewidth]{figures/eigenpaths/#1}}
\centering
\image{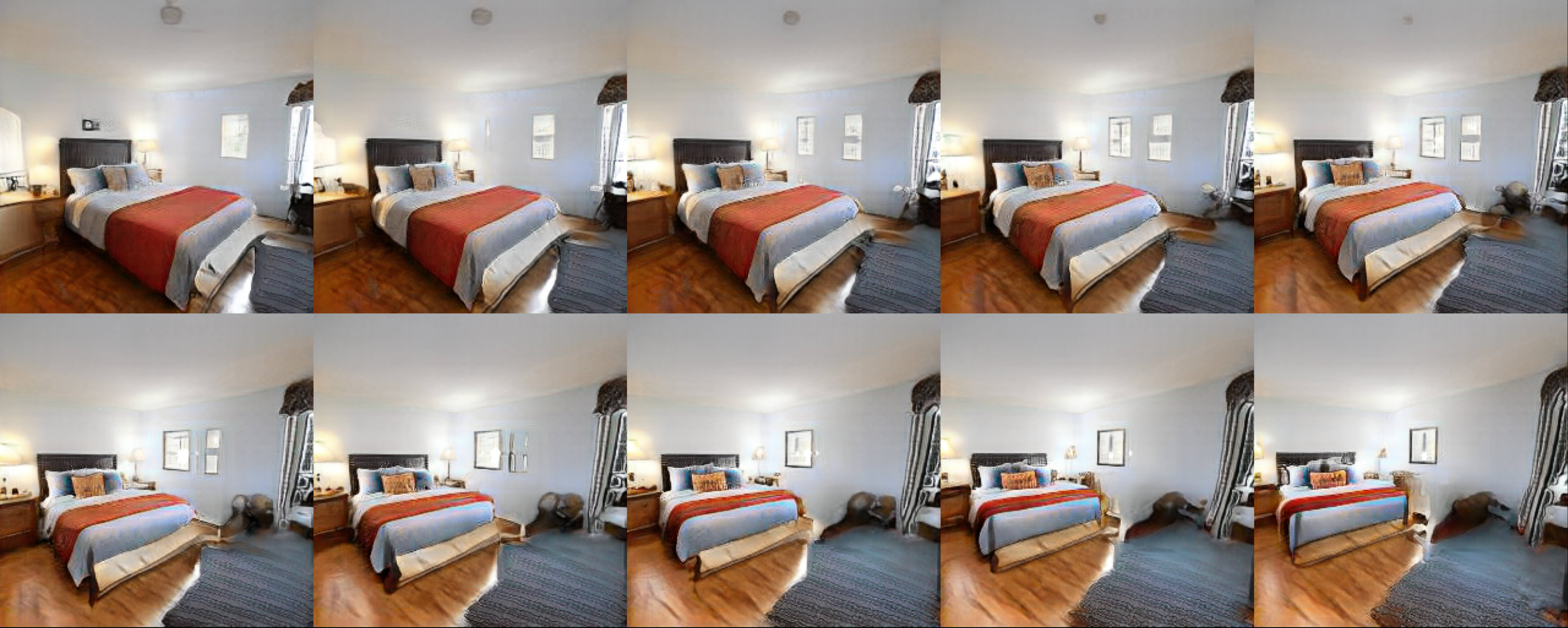}
\caption{Trajectory obtained by following the leading eigenvector of~$M_z(z_0)$ from a fixed
embedding~$z_0 \in \reals^{256}$ in the latent space of a~GAN trained on the LSUN Bedrooms dataset
at~${256 \times 256}$ resolution. The camera angle is varied smoothly, while the content of the
bedroom is preserved. The parameters used for Algorithm~\ref{alg:trace_eigenpath} are given
by~${\alpha, \rho, N} = 1.5 \cdot 10^{-2}, 0.99, 2000$; we show iterates~$0, 20, \ldots, 180$.  The
architecture and training procedure are as described in~\citep{GANConvergence}. Videos of the
trajectories for the first four eigenvectors can be viewed at this~URL:
\url{https://drive.google.com/open?id=1_SxlvcjakzkL8vNY4hporwtvg8mLQw4S}.}
\end{figure}

Learning disentangled representations is of interest, because they allow us gauge the ability of
generative models to benefit downstream tasks. For example, a robotics application may require a
vision model that reliably determines the orientation of an object, subject to changes in viewpoint,
lighting conditions, and intra-class variation in appearance. A generative model with independent
control over the full extent of each of these attributes may help to achieve the level of robustness
that is required for these applications. Moreover, the ability to synthesize concepts in ways that
are not present in the original data (e.g., rotation of an object to a position not encoutered
during training) is a useful benchmark for reasoning and out-of-distribution generalization. Often
times, the goal of training a generative model is to observe that this type of generalization
occurs.

Following~\citep{betaVAE}, we say a generative model~$G : Z \to X$, where~${Z \subset \reals^m}$
and~${X \subset \reals^n}$, possesses a disentangled representation when it satisfies two important
properties. The first property is \emph{independence} of the components of the latent variable.
Roughly speaking, this means that a perturbation to any component of a latent variable should result
in a change to the output that is distinct, in some sense, from the changes resulting from
perturbations to the other components. For example, if the first component of the latent variable
controls hair length, then the other components should not influence this attribute. The second
property is \emph{interpretability} of the changes in~$X$ resulting from these perturbations.
Suppose that the process of sampling from the distribution~$\pdata$ modeled by~$G$ can be realized
by a simulator that is parameterized over a list of independent, scalar-valued factors or
attributes. These attributes might correspond to concepts such as lighting, azimuth, gender, age,
and so on. This property is met when~$G(z + \alpha e_i)$ results in a change along exactly one of
these attributes, for each~$i \in [1, m]$, where~$e_i$ is the~$i$th standard basis vector
and~$\alpha > 0$. Quantifying the extent to which this property holds is generally challenging.

Classical dimensionality reduction techniques, such as~PCA and metric~MDS, can be used to obtain a
latent representation satisfying the first property. For instance, the former ensures that changes
to the output resulting from changes to distinct components of the latent variable are orthogonal,
which is a particularly restrictive form of distinctness. These techniques are guaranteed to
faithfully represent the underlying structure only when the data occupy a linear subspace. For many
applications of interest, such as image modeling, the data manifold could be a highly curved and
twisted surface that does not satisfy this assumption. The Swiss Roll dataset is a simple example
for which this is the case: two points with small euclidean distance could have large geodesic
distance on the data manifold. Since both techniques measure similarity between each pair of points
using euclidean distance, distant outputs in~$X$ could get mapped to nearby embeddings in~$Z$. Thus,
the second property often fails to hold in practice.

Deep generative models have the potential to learn representations that satisfy both of the required
properties for disentanglement. Our work focuses on the case in which~$G$ is a GAN generator.
Let~${\jg(z_0) \in \reals^{n \times m}}$ be the Jacobian of~$G$ evaluated at the latent
variable~${z_0 \in Z}$. Our main contributions are as follows:
\begin{enumerate}
\item We show that the leading right-singular vectors of~$\jg(z_0)$ can be used to obtain a local
disentangled representation about a neighborhood of~$z_0$;
\item We show that following the path determined by a leading right-singular vector in the latent
space of a~GAN yields a trajectory along which local changes to the output are interpretable; and
\item We formulate a regularizer that induces disentangled representations in~GANs, by aligning the
top~$k$ right-singular vectors~${v_1, \ldots, v_k}$ with the first~$k$ coordinate directions, ${e_1,
\ldots, e_k}$.
\end{enumerate}

\section{Related Work}

To date, the two most successful approaches for unsupervised learning of disentangled
representations are the $\beta$-VAE~\citep{betaVAE} and InfoGAN~\citep{InfoGAN}. The former proposes
increasing the weight~$\beta$ for the KL-divergence term in the objective of the VAE~(\citet{AEVB},
\citet{StochasticBackprop}), which is normally set to one. This weight controls the tradeoff between
minimizing reconstruction error, and minimizing the KL-divergence between the approximate posterior
and the prior of the decoder. The authors observe that as the KL-divergence is reduced, more
coordinate directions in the latent space of the decoder end up corresponding to disentangled
factors. To measure the extent to which this occurs, they describe a disentanglement metric score,
which we use in Section~\ref{sec:results}. Follow-up work~\citep{UnderstandingDisentangling}
analyzes~$\beta$-VAE from the perspective of information channel capacity, and describes principled
way of controlling the tradeoff between reconstruction error and disentanglement. Several variants
of the~$\beta$-VAE have also since been developed~(\citet{FactorVAE}, \citet{TCVAE}, \citet{HFVAE}).

InfoGAN~\citet{InfoGAN} augments the GAN objective~\citep{GAN} with a term that maximizes the mutual
information between the generated samples and small subset of latent variables. This is done by
means of an auxiliary classifier that is trained along with the generator and the discriminator.
Adversarial training offers the potential for a high degree of realism that is difficult to achieve
with models like VAEs, which are based on reconstruction error. However, InfoGAN finds fewer
disentangled factors than VAE-based approaches on datasets such as CelebA~\citep{CelebA} and
3DChairs~\citep{3DChairs}, and offers limited control over each latent factor (e.g., maximum
rotation angle for azimuth). The mutual information regularizer also detriments sample quality.
Hence, the development of an unsupervised method for learning disentangled representations with
GANs, while meeting or exceeding the quality of those found by VAE-based approaches, remains an open
problem.

Our work makes an important step in this direction. In contrast to previous approaches, which are
based on information theory, we leverage the spectral properties of the Jacobian of the generator.
This new perspective not only allows us to induce high-quality disentangled representations in~GANs,
but also to find curved paths over which disentanglement occurs, for~GANs that do not possess
disentangled representations.

\section{Identifying Local Disentangled Factors} \label{sec:local_factors}

We begin by describing how the singular value decomposition of~$\jg(z_0)$ can be used to define a
local generative model about~$z_0$ that satisfies the first property of disentangled
representations. In particular, we show that the left-singular vectors of~$\jg(z_0)$ form an
orthonormal set of directions from~$G(z_0)$ along which the magnitudes of the instantaneous changes
in~$G$ are maximized. Since perturbations from~$z_0$ along the right-singular vectors result in
changes from~$G(z_0)$ along the left-singular vectors, the right-singular vectors result in distinct
changes to the output of~$G$. This simple relationship allows us to define a local generative model
that satisfies the independence property.

First, we show that the perturbations from~$z_0$ along the right-singular vectors maximize the
magnitude of the instantaneous change in~$G(z_0)$. Given an arbitrary vector~${v \in \reals^m}$, the
directional derivative
\begin{equation}
	\lim_{\epsilon \to 0} \frac{G(z_0 + \epsilon v) - G(z_0)}{\epsilon} = \jg(z_0) v.
\end{equation}
measures the instantaneous change in~$G$ resulting from a perturbation along~$v$ from~$z_0$. The
magnitude of this change is given by
\begin{align}
	\| \jg(z_0) v \|
	&= (v^t \jg(z_0)^t \jg(z_0) v)^{1 / 2} \notag\\
	&\eqqcolon (v^t M_z(z_0) v)^{1 / 2}
	\eqqcolon n_z(v, z_0),
\end{align}
which is a seminorm involving the positive semidefinite matrix~${M_z(z_0) \coloneqq \jg(z_0)^t
\jg(z_0) \in \reals^{m \times m}}$. The unit-norm perturbation from~$z_0$ that maximizes~$n_z(v,
z_0)$ is given by
\begin{equation}
	v_1
	\coloneqq \max_{v \in \mathbb{S}^{m - 1}} n_z(v, z_0)
	= \max_{v \in \mathbb{S}^{m - 1}} v^t \jg(z_0)^t \jg(z_0) v,
\end{equation}
where~$\mathbb{S}^{m - 1} \subset \reals^m$ is the unit sphere. This is the first eigenvector
of~$M_z(z_0)$, which coincides with the first right-singular vector of~$\jg(z_0)$. It follows from
the singular value decomposition of~$\jg(z_0)$ that the first left-singular vector is given by~${u_1
= \sigma_1^{-1} \jg(z_0) v_1}$, where~$\sigma_1$ is the first singular value. Hence, a perturbation
from~$z_0$ along~$v_1$ maximizes the magnitude of the instantaneous change in~$G(z_0)$, and this
change occurs along~$u_1$.

Next, we consider the unit-norm perturbation orthogonal to~$v_1$ that maximizes~$n_z(v, z_0)$. It is
given by
\begin{equation}
	v_2 \coloneqq \max_{v \in \mathbb{S}^{d - 1} \;\cap\;
		\vspan(v_1)^\perp} v^t \jg(z_0)^t \jg(z_0) v.
\end{equation}
This is the second eigenvector of~$M_z(v, z_0)$, which coincides with the second right-singular
vector of~$\jg(z_0)$. As before, we get~${u_2 = \sigma_2^{-1} \jg(z_0) v_2}$, where~$\sigma_2$ is
the second singular value. So a perturbation from~$z_0$ along~$v_2$ results in an instantaneous
change in~$G(z_0)$ along~$u_2$. Continuing in this way, we consider the~$k$th unit-norm perturbation
orthogonal to~${v_1, \ldots, v_{k - 1}}$ that maximizes~$n_z(v, z_0)$, for each~$k \in [2, r]$,
where~$r \coloneqq \rank(M_z(z_0))$. This shows that the right-singular vectors of~$\jg(z_0)$
maximize the magnitude of the instantaneous change in magnitude of~$G(z_0)$, and these changes occur
along the corresponding left-singular vectors.

Now, we use the right-singular vectors of~$\jg(z_0)$ to define a local generative model about~$z_0$.
Consider the function~${\bar{G}_{z_0} : \reals^r \to \reals^n}$,
\[
	\bar{G}_{z_0} : \alpha \mapsto G\left( z_0 + \sum_{i \in [1, r]} \alpha_i v_i \right).
\]
The components of~$\alpha$ control perturbations along orthonormal directions, and these directions
also result in orthonormal changes to~$G(z_0)$. Hence, $\bar{G}_{z_0}$~satisfies the first property
for a generative model to possess a disentangled representation, but only about a neighborhood
of~$z_0$. Figure~\ref{fig:latent_perturbations} in Appendix~\ref{sec:supplementary_figures}
investigates whether~$\bar{G}_{z_0}$ also satisfies the second property: interpretability of changes
to individual components of~$\alpha$. We can see that perturbations along the leading
eigenvectors~$M_z(z_0)$, especially the principal eigenvector, often result in the most drastic
changes. These changes are interpretable, and tend to make modifications to isolated attributes of
the face. To see this in more detail, we consider the top two rows of subfigure~(g). Movement along
the first two eigenvectors changes hair length and facial orientation; movement along the third
eigenvector decreases the length of the bangs; movement along the fourth and fifth eigenvectors
changes background color; and movement along the sixth and seventh eigenvectors changes hair color.

\section{Finding Quasi-Disentangled Paths} \label{sec:finding_paths}

\begin{figure}[t]
\newcommand{\image}[2]{\includegraphics[width=#1\textwidth]{figures/orbits/dsprites/#2}}
\newcommand{\imageA}[2]{\includegraphics[width=#1\textwidth]{figures/orbits/dsprites_aligned/#2}}
\centering
\begin{tabular}[t]{c@{\hskip 1em}c}%
\subfloat[]{%
	\image{0.2}{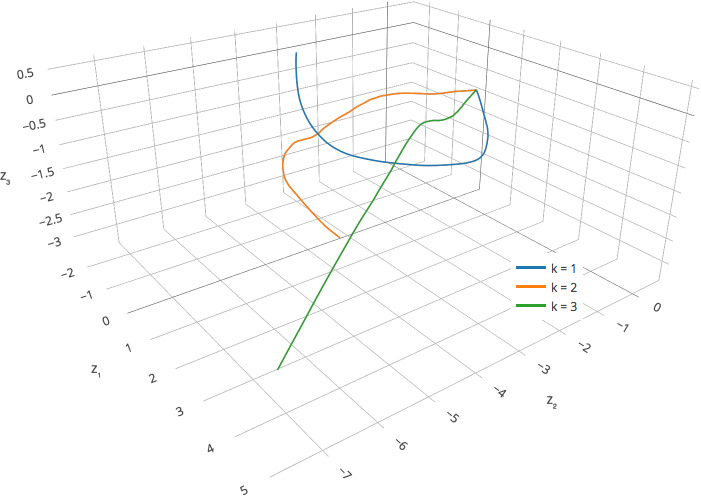}%
} &
\captionsetup{width=.4\linewidth}%
\subfloat[]{%
	\raisebox{0.13in}{%
	{\def\arraystretch{0}%
	\begin{tabular}[b]{c}%
	\image{0.2}{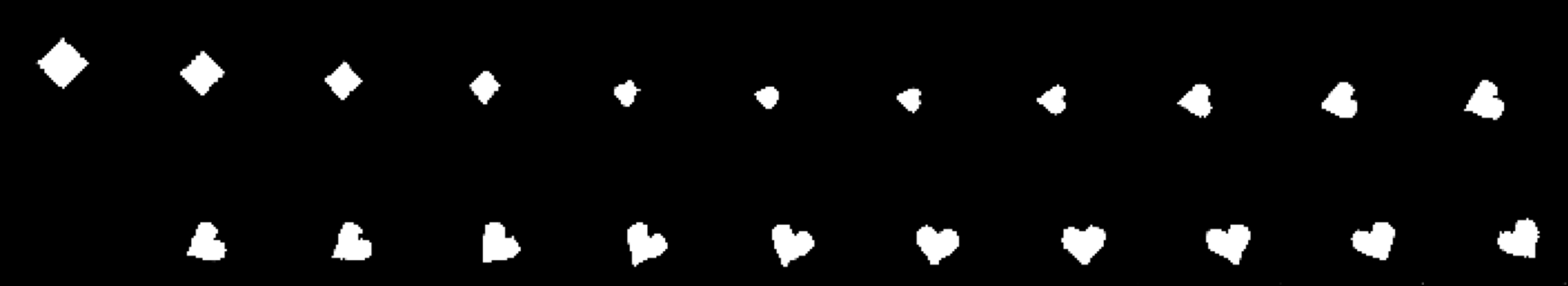} \\
	\image{0.2}{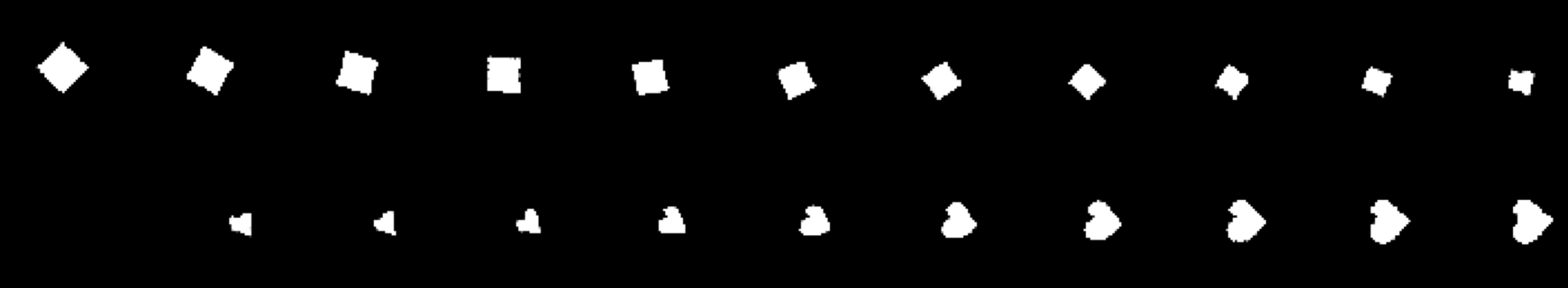} \\
	\image{0.2}{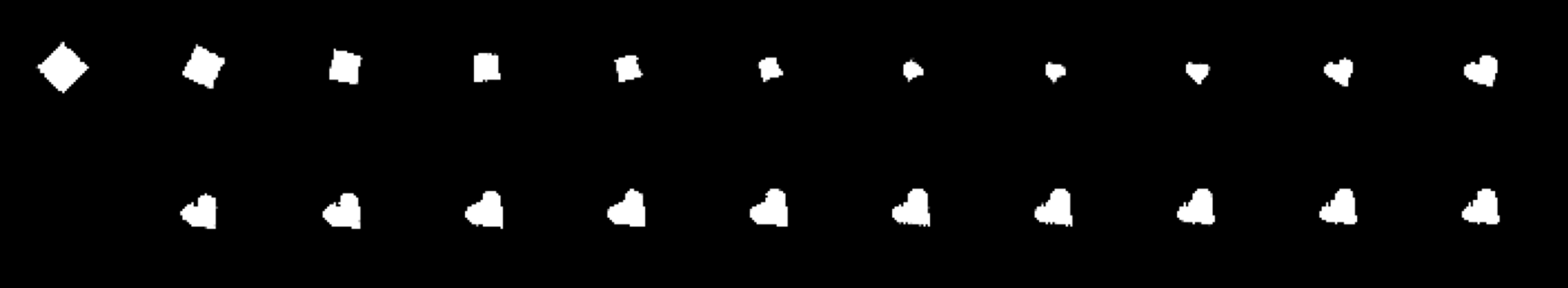}
	\end{tabular}}}%
} \\
\captionsetup{width=.4\linewidth}%
\subfloat[]{%
	\imageA{0.2}{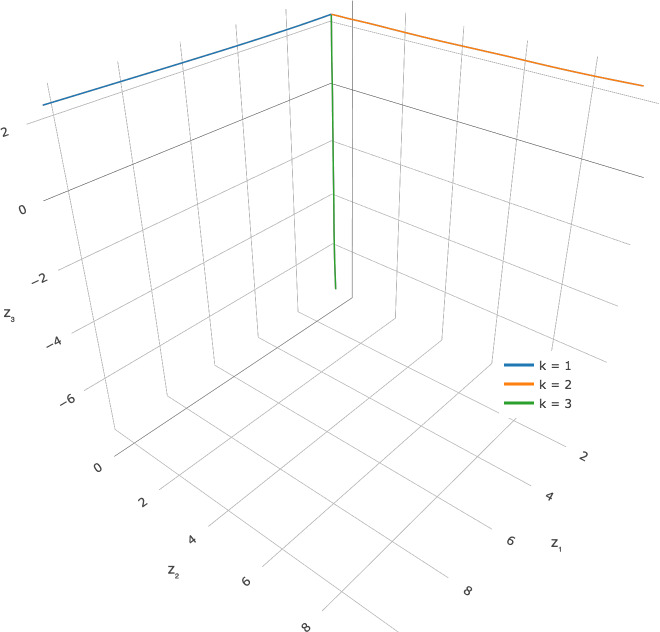}%
} &
\captionsetup{width=.4\linewidth}%
\subfloat[]{%
	\raisebox{0.28in}{%
	{\def\arraystretch{0}%
	\begin{tabular}[b]{c}%
	\imageA{0.2}{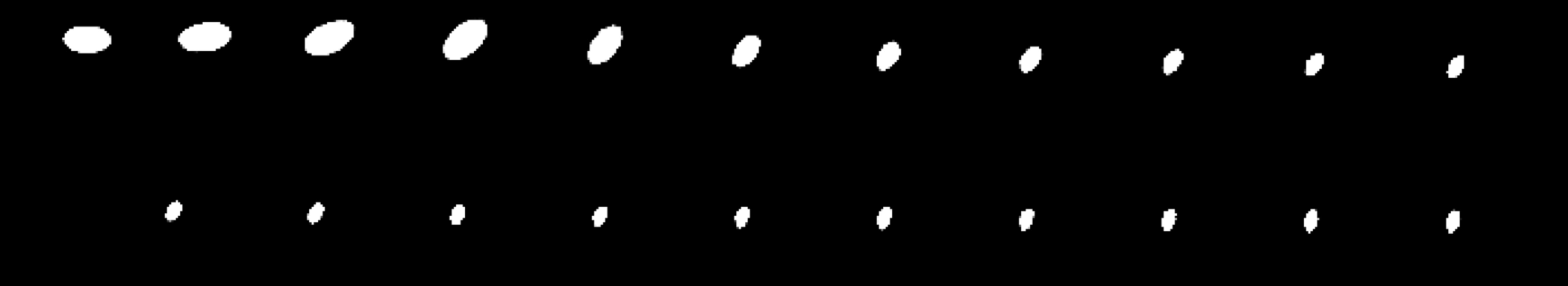} \\
	\imageA{0.2}{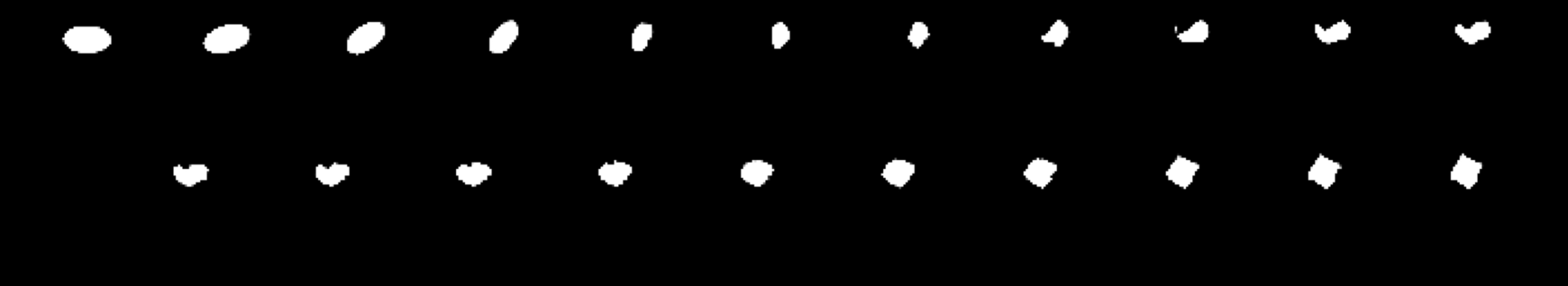} \\
	\imageA{0.2}{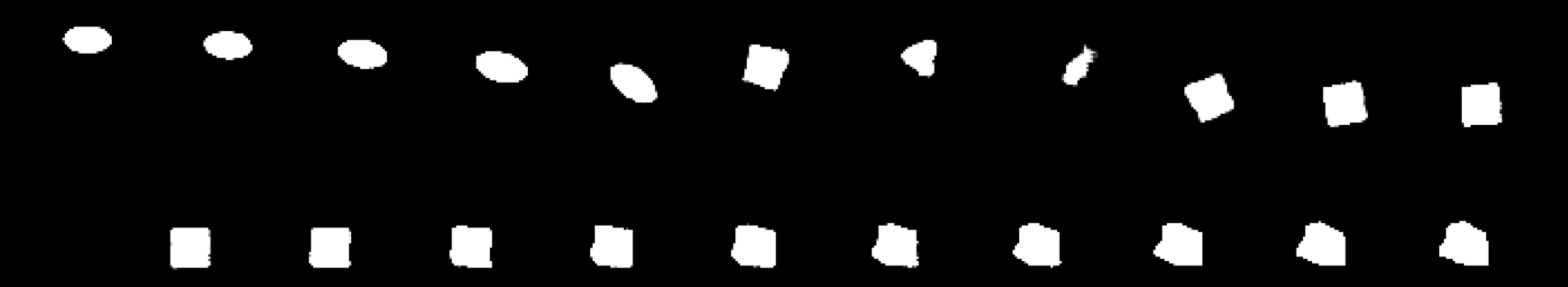}
	\end{tabular}}}%
} \\
\end{tabular}
\caption{Trajectories obtained by following the first three eigenvectors of~$M_z(z_0)$ from a fixed
embedding~$z_0 \in \reals^3$ in the latent spaces of two~GANs with identical architectures, trained
on the dSprites dataset~\citep{dSprites}. Subfigure~(a) shows plots of the
trajectories~${\{\gamma_k\}_{k \in [1, 3]}}$, and subfigure~(b) shows the outputs of~$G$ at iterates~$0,
100, \ldots, 2000$ of each trajectory~$\gamma_k$, for~$k \in [1, 3]$, from top to bottom,
respectively. Subfigures~(c) and~(d) show the same information for an alignment-regularized
GAN~(${k, \lambda = 3, 0.1}$); the trajectories are now axis-aligned, as expected. The parameters
used for Algorithm~\ref{alg:trace_eigenpath} are given by~${\alpha, \rho, N} = 5 \cdot 10^{-3},
0.99, 2000$. Details regarding the model architecture are given in
Appendix~\ref{sec:model_arch}.\label{fig:eigenpaths_dsprites}}
\end{figure}

Generative models known to possess disentangled representations, such
as~$\beta$-VAEs~\citep{betaVAE}, allow for continuous manipulation of attributes via perturbations
to individual coordinates. Starting from a latent variable~$z_0 \in Z$, we can move along the
path~${\gamma : t \mapsto z_0 + t e_i}$ in order to vary a single attribute of~$G(z_0)$, while
keeping the others held fixed. GANs are not known to learn disentangled representations
automatically. Nonetheless, the previous section shows that the local generative
model~$\bar{G}_{z_0}$ does possess a disentangled representation, but only about a neighborhood of
the base point~$z_0$. We explore whether it is possible to extend this local model to obtain
disentanglement along a continuous path from~$z_0$. To do this, we construct a trajectory~${\gamma_k
: \reals \to \reals^n}$, ${t \mapsto G(\gamma_k(t))}$ by repeatedly following the~$k$th leading
eigenvector of~$M_z(\gamma_k(t))$, where~$\gamma(0) \coloneqq z_0$. The procedure used to do this is
given by Algorithm~\ref{alg:trace_eigenpath}.

We first test the procedure on a toy example for which it is possible to explicitly plot the
trajectories. We use the dSprites dataset~\citep{dSprites}, which consists of~$64 \times 64$ white
shapes on black backgrounds. Each shape is completely determined by six attributes: symbol~(square,
ellipse, or heart), scale~(6 values), rotation angle from~$0$ to~$2 \pi$~(40 values), and $x$-~and
$y$-positions~(30 values each). We trained a~GAN on this dataset using a latent variable size of
three, fewer than the six latent factors that determine each shape.
Figure~\ref{fig:eigenpaths_dsprites} shows that outputs of the generator along these trajectories
vary locally along only one or two attributes at a time. Along the first trajectory~$\gamma_1$, the
generator first decreases the scale of the square, then morphs it into a heart, increases the scale
again, and finally begins to rotate it. Similar comments apply to the other two trajectories,
$\gamma_2$ and~$\gamma_3$.

\begin{figure}[t]
\newcommand{\fullwidthimage}[1]{\includegraphics[width=0.5\textwidth]{figures/orbits/#1}}
\centering
\subfloat[][Trajectories corresponding to the principal eigenvector for embeddings~$z_1, \ldots,
z_4$, from top to bottom, respectively. For embeddings~$z_1, z_2$, and~$z_3$, we show iterates~$0,
20, \ldots, 180$, and for embedding~$z_4$, we show iterates~$0, 10, \ldots, 90$. The first
trajectory varies azimuth; the second, gender; the third, hair length; and the fourth, hair
color.]{%
\fullwidthimage{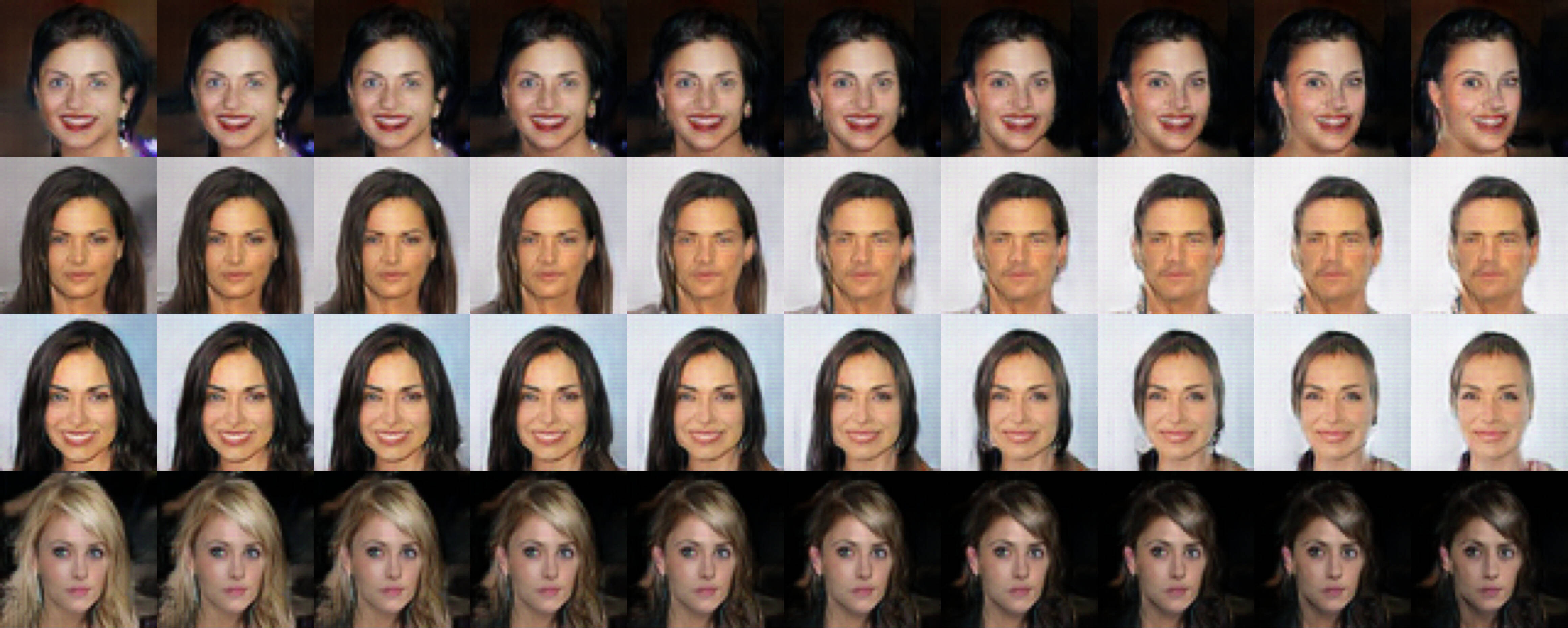}} \\
\subfloat[][Trajectories corresponding to the first, second, third, and fifth leading eigenvectors
of embedding~$z_5$, from top to bottom, respectively; we show iterates~$0, 20, \ldots, 200$. The
first trajectory primarily varies azimuth; the second, gender; the third, age; and the fourth, hair
color and presence of facial hair.]{%
\fullwidthimage{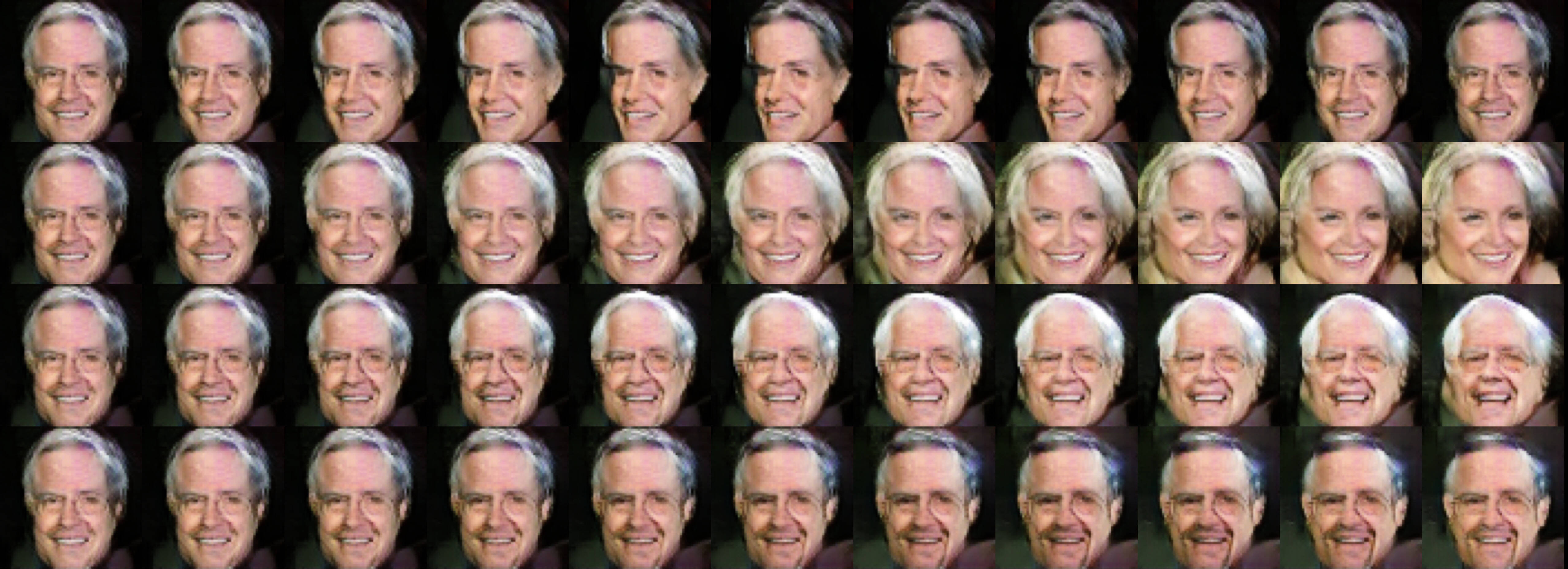}}%
\caption{Trajectories found by following leading eigenvectors from five fixed embeddings~$z_1,
\ldots, z_5$, using Algorithm~\ref{alg:trace_eigenpath} (${\alpha = 1.5 \cdot 10^{-2}}, {\rho =
0}$). Details regarding the model architecture are given in
Appendix~\ref{sec:model_arch}.\label{fig:eigenpaths_celeb_a}}
\end{figure}

Next, we test the procedure on a~GAN trained on the CelebA~dataset~\citep{CelebA} at~$64 \times 64$
resolution. Figure~\ref{fig:eigenpaths_celeb_a}(a) shows the trajectories~$\gamma_1$ starting at
four fixed embeddings~${z_1, \ldots, z_4}$. Although the association between the ordinal~$k$ of the
eigenvector~$v_k$ and the attribute of the image being varied is not consistent throughout latent
space, local changes still tend to occur along only one or two attributes at a time.
Figure~\ref{fig:eigenpaths_celeb_a}(b) shows the trajectories~${\gamma_1, \gamma_2, \gamma_3}$
and~$\gamma_5$, all starting from the same fixed embedding~$z_5$. As is the case for the~dSprites
dataset, we can see that trajectories~$\gamma_k$ for distinct~$k$ tend to effect changes to~$G(z_0)$
along distinct attributes. These results suggest that, along trajectories from~$z_0$ determined by a
leading eigenvector of~$M_z(z_0)$, changes in the output tend to occur along isolated attributes.

\begin{algorithm}[t]\tiny
\caption{Procedure to trace path determined by $k$th~leading
eigenvector.\label{alg:trace_eigenpath}}
\newcommand{\mv}{\operatorname{mv}}

\begin{algorithmic}[1]
\Require{${\mv : \reals^m \times \reals^m \to \reals^d}$, ${z, v \mapsto M_z(z) v}$ is a function
that computes matrix-vector products with the implicitly-defined matrix~$M_z(z) \in \reals^{m \times
m}$.}
\Require{$z \in \reals^m$ is the embedding from which to begin the trajectory.}
\Require{$k \in [1, m]$ is ordinal of the eigenvector to trace, with~$k = 1$ corresponding to the
leading eigenvector.}
\Require{$\alpha > 0$ is the step size.}
\Require{$\rho \in [0, 1)$ is the decay factor.}
\Require{$N \geq 1$ is the required number of steps in the trajectory.}
\State

\Procedure{TraceEigenpath}{$\mv, z, k, \alpha, \rho, N$}
\State $z_0 \gets z$
\For{$i \in [1, N]$}
	\State $M_z(z) \gets \Call{EvaluateNormalJacobian}{\mv, z}$ \Comment{Details in Appendix~\ref{sec:autodiff}}
	\State $V, D, V^t \gets \Call{SVD}{M_z(z)}$
	\State $w_i \gets v_k$ \Comment{Take the~$k$th eigenvector}
	\If{$i \geq 2$}
		\If{$\inner{w_{i - 1}}{w_i} < 0$}
			\State $w_i \gets -w_i$ \Comment{Prevent backtracking by ensuring that
			${\angle(w_{i - 1}, w_i) \leq \pi / 2}$}
		\EndIf
		\State $w_i \gets \rho w_{i - 1} + (1 - \rho) w_i$ \Comment{Apply decay to smoothen
		the trajectory}
	\EndIf
	\State $z_i \gets z_{i - 1} - \alpha w_i$
\EndFor
\State \Return $\{z_0, \ldots, z_N\}$
\EndProcedure
\end{algorithmic}
\end{algorithm}

\section{Aligning the Local Disentangled Factors} \label{sec:aligning_local_factors}

\begin{figure}[t]
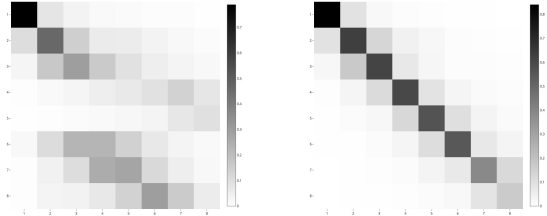

\centering
\newcommand{\image}[1]{\includegraphics[width=0.20\textwidth]{figures/heatmaps/#1}}
\subfloat{\image{{gan_nz_128_ac_8_aw_0.1_nw}.jpg}}
\hspace{0.2in}
\subfloat{\image{{gan_nz_128_ac_8_aw_0.1}.jpg}}
\caption{Comparison of the top-left $8 \times 8$~corner of the
matrix~$F$~(Equation~\ref{eq:heatmap_matrix}) for alignment-regularized GANs~(${k = 8, T = 8,
\lambda = 0.1}$), trained with reweighting~(left) and without reweighting~(right) of the columns of
the matrix~$S_k$ defined by Equation~\ref{eq:reg_mask}. Both GANs were trained on the CelebA
dataset~\citep{CelebA} at~$64 \times 64$ resolution, with latent variable size~128. See
Appendix~\ref{sec:model_arch} for complete details regarding the model architecture and training
procedure.\label{fig:alignment_heatmaps}}
\end{figure}

The~$\beta$-VAE~\citep{betaVAE} is known to learn disentangled representations, so that traveling
along paths of the form
\begin{equation}
	\gamma : t \mapsto z_0 + t e_j,
	\label{eq:aligned_path}
\end{equation}
for certain coordinate directions~$e_j$, produces changes to isolated attributes of~$G(z_0)$. In the
previous section, we saw that such paths still exist for GANs, but they are not simply straight
lines oriented along the coordinate axes~(see Figure~\ref{fig:eigenpaths_dsprites}). We develop an
efficient regularizer that encourages these paths to take the form of
Equation~\ref{eq:aligned_path}, based on alignment of the top~$k$ eigenvectors of~$M_z(z_0)$ with
the first~$k$ coordinate directions~${e_1, \ldots, e_k}$. Before proceeding, we describe a useful
visualization technique to help measure the extent to which this happens. Let~${M_z(z_0) = V D V^t}$
be the eigendecomposition of~$M_z(z_0)$, where~$V$ is an orthogonal matrix whose columns are
eigenvectors, and~$D$ the diagonal matrix of nonnegative eigenvalues, sorted in descending order.
Now we define~${\tilde{V} : z \mapsto V}$ to be the function that maps~$z$ to the corresponding
eigenvector matrix~$V$, and let
\begin{equation}
	F \coloneqq \E_{z \sim p_z} \tilde{V}(z) \circ \tilde{V}(z),
	\label{eq:heatmap_matrix}
\end{equation}
where~$p_z$ is the prior over~$Z$, and~`$\circ$' denotes Hadamard product. If the~$k$th column is
close to a one-hot vector, with values close to zero everywhere except at entry~$j$, then we know
that on average, the~$k$th leading eigenvector of~$M_z(z)$ is aligned with~$e_j$. A heatmap
generated from this matrix therefore allows us to gauge the extent to which each eigenvector~$v_k$
is aligned with the coordinate direction~$e_k$. Figure~\ref{fig:alignment_heatmaps} shows that this
does not happen automatically for a~GAN, even when it is trained with a small latent variable size.
Interestingly, eigenvector alignment does occur for~$\beta$-VAEs.
Appendix~\ref{sec:beta_vae_alignment} explores this connection in more detail.

We begin by considering the case where we only seek to align the leading eigenvector~${v_1 \in
\reals^m}$ with the first coordinate direction~${e_1 \in \reals^m}$. A simple way to do this is to
obtain an estimate~$\hat{v}_1$ for~$v_1$ using~$T$ power iterations, and then
renormalize~${\hat{v}_1 \in \reals^m}$ to a unit vector. We can then maximize the value of the first
component of the elementwise squared vector~${\hat{v}_1 \circ \hat{v}_1}$, and minimize values of
the remaining components. Using the mask~${s_1 \coloneqq (-1, 1, \ldots, 1) \in \reals^m}$, we
define the regularizer~${R_1 : \reals^m \to \reals}$,
\begin{equation}
	R_1(z) \coloneqq \sum_{i \in [1, m]} (s_1 \circ \hat{v}_1 \circ \hat{v}_1)_i.
\end{equation}
Since~$\hat{v}_1$ is constrained to unit norm, this regularizer is bounded. It can be incorporated
into the loss for the generator using an appropriate penalty weight~${\lambda > 0}$.

Next, we consider the case where we would like to align the first two leading eigenvectors~${v_1,
v_2 \in \reals^m}$ with the first two coordinate directions~${e_1, e_2 \in \reals^m}$. One potential
approach is to first compute an estimate~$\hat{v}_1$ to~$v_1$ using~$T$ power iterations, as before.
Then, we could apply a modified version of the power method to obtain an estimate~$\hat{v}_2$
for~$v_2$, in which we project the result of each power iteration onto~$\vspan(\hat{v}_1)^\perp$
using the projection~${P_1 \coloneqq I - \hat{v}_1 \hat{v}_1^t}$. There are two problems with this
approach. Firstly, it could be inaccurate: unless~$\|v_1 - \hat{v}_1\| < \tau$ for sufficiently
small~$\tau > 0$, which may require a large number of power iterations, $P_1$~will not be an
accurate projection onto~$\vspan(v_1)^\perp$. Error in approximating~$v_1$ would then jeopardize the
approximation to~$v_2$. Second, the approach is inefficient. We can only run the power method to
estimate~$v_2$ after the we have already obtained an estimate for~$v_1$. If we use~$T$ power
iterations to estimate each eigenvector, then estimating the first~$k$ eigenvectors will require a
total of~$kT$ power iterations. This is too slow to be practical.

Our specific application of the power method enables an optimization that allows for the~$k$ power
iterations to be run simultaneously. Once we apply the regularizer to the GAN~training procedure,
the first eigenvector~$v_1$ will quickly align with the first coordinate direction~$e_1$. We
therefore assume that~${v_1 = e_1}$. This assumption would imply that ${\vspan(v_1)^\perp =
\vspan(e_2, \ldots, e_m)}$, so applying~$P_1$ would amount to zeroing out the first component
of~$\hat{v}_2$ after each power iteration. Since~$P_1$ would no longer depend on~$\hat{v}_1$, we can
run the power iterations for~$v_1$ and~$v_2$ in parallel. To formally describe this, we let~${c_p
\coloneqq 1 \in \reals^p}$ be the constant vector of ones, and let
\begin{equation}
	M_2 \coloneqq \begin{pmatrix*} c_m & \begin{matrix*} 0 \\ c_{m - 1} \end{matrix*} \end{pmatrix*}.
	\label{eq:eigvec_mask_2}
\end{equation}
Given a matrix~${\hat{V}_t^{(2)} \in \reals^{m \times 2}}$ whose columns are the current estimates
for~$v_1$ and~$v_2$, respectively, we can describe the power iterations for~$v_1$ and~$v_2$ using
the recurrence
\begin{equation}
	\hat{V}_{t + 1}^{(2)} \coloneqq M_2 \circ (M_z(z_0) \hat{V}_t^{(2)}).
	\label{eq:power_iter_2}
\end{equation}
Now, let~$\hat{V}^{(2)} \in \reals^{m \times 2}$ be the final estimate for~$v_1$ and~$v_2$. To
implement the regularizer, we let~$s_2 \coloneqq (1, -1, 1, \ldots, 1) \in \reals^m$, $S_2 \coloneqq
\begin{pmatrix*} s_1 & s_2 \end{pmatrix*} \in \reals^{m \times 2}$, and define
\begin{equation}
	R_2(z) \coloneqq \sum_{\substack{i \in [1, m] \\ j \in [1, 2]}}
		(S_2 \circ \hat{V}^{(2)} \circ \hat{V}^{(2)})_{i, j}.
	\label{eq:reg_2}
\end{equation}

\begin{algorithm}[t]\scriptsize
\caption{Procedure to estimate the top~$k$ eigenpairs of~$M_z(z)$.\label{alg:est_dom_eigenpairs}}
\newcommand{\mv}{\operatorname{mv}}

\begin{algorithmic}[1]
\Require{$\mv : \reals^m \to \reals^m$ is a function that computes matrix-vector products with the
implicitly-defined matrix~$M_z(z) \in \reals^{m \times m}$.}
\Require{$V \in \reals^{m \times k}$ is a matrix whose columns are the initial estimates for the
eigenvectors.}
\Require{$T \geq 1$ is the required number of power iterations.}
\State

\State $\epsilon \gets 10^{-8}$ \Comment{Guards against division by numbers close to zero}
\State

\Procedure{EstimateLeadingEigenpairs}{$\mv, V, T$}
\State Let~$M_k \in \reals^{m \times k}$ be given by Equation~\ref{eq:eigvec_mask}
\State $V_0 \gets M_k \circ V$ \Comment{`$\circ$' denotes Hadamard product}
\State

\For{$i \in [1, T]$}
	\State $V_i \gets M \circ \mv(V_{i - 1})$
	\State $\Lambda_i \gets \diag(\Call{ColumnNorms}{V_i})$
	\State $V_i \gets V_i\, (\Lambda_i + \epsilon I)^{-1}$ \Comment{Renormalize columns}
\EndFor

\State \Return $\Lambda_T, V_T$
\EndProcedure
\State

\Procedure{ColumnNorms}{$A$} \Comment{$A \in \reals^{p \times q}$}
	\State \Return $(\|a_1\|, \ldots, \|a_q\|)$ \Comment{$a_i \in \reals^p$ is the $i$th column
	of $A$}
\EndProcedure
\end{algorithmic}
\end{algorithm}

\begin{algorithm}[t]\scriptsize
\caption{Procedure to evaluate the alignment penalty.\label{alg:alignment_reg}}
\newcommand{\mv}{\operatorname{mv}}

\begin{algorithmic}[1]
\Require{$k \in [1, m]$ is number of leading eigenvectors to align with~$e_1, \ldots, e_k$.}
\Require{$\mv, T$ are as defined in Algorithm~\ref{alg:est_dom_eigenpairs}.}
\State

\Procedure{EvaluateAlignmentRegularizer}{$k, \mv, T$}
\State Let~$S_k$ be given by Equation~\ref{eq:reg_mask}
\State

\State $\alpha \gets 2 / (k (k + 1))$
\State $A \gets \diag(\alpha \cdot (k, k - 1, \ldots, 1))$
\State $S_k \gets S_k A$ \Comment{Reweight columns to prioritize alignment of leading eigenvectors}
\State

\State $V_0 \gets \Call{RandomRademacher}{m, k}$
\State $\hat{\Lambda}, \hat{V} \gets \Call{EstimateLeadingEigenpairs}{\mv, V_0, t}$
\State \Return $\Call{Sum}{S_k \circ \hat{V} \circ \hat{V}}$
\State
\EndProcedure

\Procedure{RandomRademacher}{$p, q$}
\State \Return $A \in \reals^{p \times q}$, where~$a_{ij} = 1$ with probability~$1 / 2$ and~$-1$
with probability~$1 / 2$
\EndProcedure
\end{algorithmic}
\end{algorithm}

\begin{figure}[t]
\centering
\newcommand{\image}[1]{\includegraphics[width=0.9\linewidth]{figures/alignment_penalty/#1}}
\hspace{-0.15in}\image{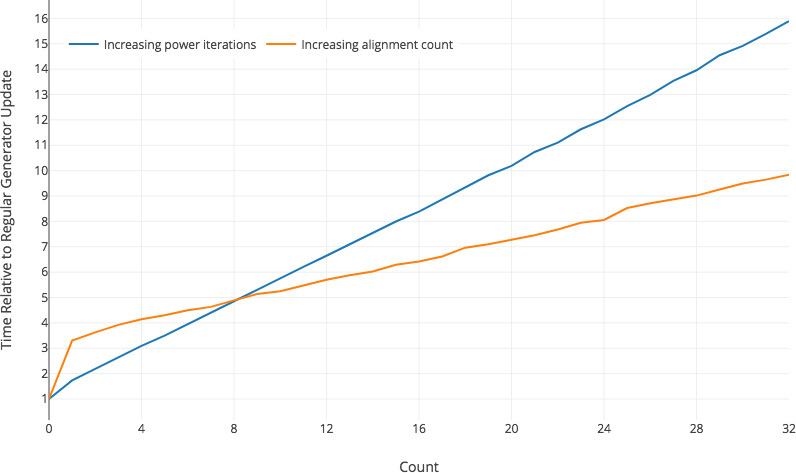}
\caption{Median relative cost per generator update as a function of the number of eigenvectors~$k$
to align, and the number of power iterations~$T$ to use for Algorithm~\ref{alg:alignment_reg}. The
cost is measured relative to the time required to make one RMSProp~\citep{RMSProp} update to the
parameters of a~DCGAN generator~(approximately $20.419$~ms) with base feature map count~$64$ and
batch size~$32$. Complete details regarding the model architecture are given in
Appendix~\ref{sec:model_arch}. Medians were computed using the update times for the first~$500$
iterations; the median absolute deviations are small enough as to be indistinguishable from the
medians on the plot.\label{fig:alignment_reg_perf}}
\end{figure}

It is straightforward to generalize this approach to the case where we seek to align the first~$k$
eigenvectors~${v_1, \ldots, v_k}$ with~${e_1, \ldots, e_k}$. For each eigenvector~$v_i$, with~${i
\in [2, k]}$, we assume that eigenvectors~${v_1, \ldots v_{i - 1}}$ are already aligned with~${e_1,
\ldots, e_{i - 1}}$. The projections onto~$\vspan(e_1)^\perp, \,\vspan(e_1, e_2)^\perp, \ldots,
\,\vspan(e_1, \ldots, e_{i - 1})^\perp$ can be implemented using columns~${2, 3, \ldots, i}$,
respectively, of the mask~$M_k \in \reals^{m \times k}$. This mask, which is a generalization of the
one defined by Equation~\ref{eq:eigvec_mask_2}, is given by~${\reals^{m \times k} \ni M_k \coloneqq}$
\begin{equation}
	\smallmath{\begin{pmatrix*}
	c_k & c_k - e_1 & c_k - (e_1 + e_2) & \cdots & c_k - (e_1 + \cdots + e_{k - 1}) \\
	c_{m - k} & c_{m - k} & c_{m - k} & \cdots & c_{m - k}
	\end{pmatrix*}}. \label{eq:eigvec_mask}
\end{equation}
The resulting procedure to estimate the leading~$k$ eigenvectors is described by
Algorithm~\ref{alg:est_dom_eigenpairs}. Figure~\ref{fig:alignment_reg_perf} shows that the runtime
of Algorithm~\ref{alg:est_dom_eigenpairs} scales linearly with respect to the number of
eigenvectors~$k$ and the number of power iterations~$T$. Next, we generalize
Equation~\ref{eq:reg_2}, in order to describe how to evaluate the regularizer~$R_k$. Let~${s_p \in
\reals^m}$ be given by~$(s_p)_i = -1$ if~$i = p$ and~$1$ otherwise, and define
\begin{equation}
	S_k \coloneqq \smallmath{\begin{pmatrix*}[l]
	-1                & \phantom{-}1      & \phantom{-}1      & \cdots & \phantom{-}1      \\
	\phantom{-}1      & -1                & \phantom{-}1      & \cdots & \phantom{-}1      \\
	\phantom{-}1      & \phantom{-}1      & -1                & \cdots & \phantom{-}1      \\
	\phantom{-}\vdots & \phantom{-}\vdots & \phantom{-}\vdots & \ddots & \phantom{-}\vdots \\
	\phantom{-}1      & \phantom{-}1      & \phantom{-}1      & \cdots & -1                \\
	\phantom{-}\vdots & \phantom{-}\vdots & \phantom{-}\vdots & \ddots & \phantom{-}\vdots \\
	\phantom{-}1      & \phantom{-}1      & \phantom{-}1      & \cdots & \phantom{-}1
	\end{pmatrix*}} 
	= \begin{pmatrix*} s_1 & \cdots & s_k \end{pmatrix*} \in \reals^{m \times k}.
	\label{eq:reg_mask}
\end{equation}
Algorithm~\ref{alg:alignment_reg} shows how~$S_k$ is used with the result of
Algorithm~\ref{alg:est_dom_eigenpairs} to evaluate~$R_k$. Finally,
Figure~\ref{fig:eigenpaths_dsprites}(c) shows that incorporating the alignment regularizer into the
generator loss successfully aligns the trajectories produced by Algorithm~\ref{alg:trace_eigenpath}
with the coordinate axes.

If the alignment regularizer is implemented exactly as described by Equation~\ref{eq:reg_2}, it will
fail to have the intended effect. The reason for this has to do with the assumption behind the
optimization used to run the~$k$ power iterations in parallel. Before attempting to align~$v_i$
with~$e_i$, we assume that that~${v_1, \ldots, v_{i - 1}}$ are already aligned with~${e_1, \ldots,
e_{i - 1}}$. When this assumption fails to hold, the projections computed using the columns of~$M_k$
will no longer be valid. Figure~\ref{fig:alignment_heatmaps}(a) shows that the matrix~$F$ does not
have a diagonal in its top-left corner, which is what we would expect to see if~${v_1, \ldots, v_k}$
were aligned with~${e_1, \ldots, e_k}$. Fortunately, there is a simple fix that remedies the
situation. We would like to encourage the optimizer to prioritize alignment of~$v_i$ with~$e_i$ over
alignment of~${v_{i + 1}, \ldots, v_k}$ with~${e_{i + 1}, \ldots, e_k}$, for all~$i \in [1, k - 1]$.
A simple way to do this is to multiply the~$i$th column of~$M_k$ by a weight of~${(k - i + 1)
\alpha}$. We choose~$\alpha$ based on the condition that these weights sum to one, i.e.,
\begin{equation}
	\sum_{i \in [1, k]} i \alpha = 1,
	\quad\text{implying that}\quad
	\alpha = \frac{2}{k (k + 1)}.
\end{equation}
This reweighting scheme is implemented in lines~5--7 of Algorithm~\ref{alg:alignment_reg}.
Figure~\ref{fig:alignment_heatmaps}(b) confirms that this modification induces the desired structure
in the top-left corner of~$F$.

\section{Results} \label{sec:results}

\begin{table}
\centering
\renewrobustcmd{\bfseries}{\fontseries{b}\selectfont}
\renewrobustcmd{\boldmath}{}
\newrobustcmd{\B}{\bfseries}

\begin{tabular}{
l
@{\hspace{2em}}
S[table-format=3.2,mode=text,detect-weight]
@{${}\pm{}$}
S[table-format=2.1,mode=text,table-alignment=left,detect-weight]
}
\toprule
Model & \multicolumn{2}{@{\hspace{-1em}}c}{Disentanglement Score} \\
\midrule
Ground truth    & \multicolumn{2}{@{\hspace{-0.01em}}l}{100} \\
Raw pixels      & 45.75 & 0.8  \\
PCA             & 84.90 & 0.4  \\
ICA             & 42.03 & 10.6 \\
\cmidrule{1-3}
DC-IGN          & \B 99.3 & \B 0.1  \\
InfoGAN$^1$     & 73.5  & 0.9  \\
\cmidrule{1-3}
VAE (untrained) & 44.14 & 2.5  \\
VAE             & 61.58 & 0.5  \\
$\beta$-VAE     & \B 99.23 & \B 0.1  \\
\cmidrule{1-3}
Ours            & \B 92.34 & \B 0.4  \\
\bottomrule
\end{tabular}

\caption{Comparison of disentanglement metric scores for our method to those reported
in~\cite{betaVAE}. We use an alignment-regularized GAN~(${k, \lambda = 6, 0.6}$) with latent
variable size~$10$. Details regarding the model architecture and training procedure are given in
Appendix~\ref{sec:model_arch}.\label{table:disentanglement_score_results}}
\end{table}

We first make a quantitative comparison between our approach and previous methods that have been
used to obtain disentangled representations. This requires us to measure the extent to which the
second property of disentangled representations -- namely, interpretability of changes resulting
from perturbations to individual coordinates of a latent variable -- holds. Suppose that we had
knowledge of the ground truth latent factors for the dataset, along with a simulator that can
synthesize new outputs given assignments to these latent factors. Then we could generate pairs of
outputs, such that the outputs in each pair differ only along a single attribute. Let~${(x_0, x_1)}$
be one such pair, and~${z_0 \coloneqq G^{-1}(x_0)}$ and~${z_1 \coloneqq G^{-1}(x_1)}$ the
corresponding latent variables for the generator~${G : \reals^m \to \reals^n}$. If~$G$ satisfies the
second property, then we would expect~$z_0$ and~$z_1$ to be approximately equal along all
components, except the one corresponding to the attribute that was varied. Hence, ${|z_0 -
z_1|}$~should be a one-hot vector in expectation. \cite{betaVAE}~propose an evaluation metric based
on this idea. It involves training a linear classifier to predict the index of the latent factor
that was varied in order to generate each pair. At each step of training for the classifier, we
sample a batch of input-target pairs in accordance with the procedure described in~\citep{betaVAE},
and update the classifier using the cross-entropy loss.

Application of the evaluation metric to~GANs is complicated by the fact that a direct procedure to
invert the generator is usually not available. Models such as the~$\beta$-VAE consist of an encoder
that effectively functions as an inverse for the generative model, so this is not an issue. For the
purpose of evaluation, we also train an encoder to invert the fixed generator, after GAN training
has finished. This additional training procedure uses a standard autoencoding loss, and the details
are specified in Appendix~\ref{sec:disentanglement_metric_score}.
Table~\ref{table:disentanglement_score_results} compares our approach to the ones evaluated
in~\citep{betaVAE} on the dSprites dataset~\citep{dSprites}, which we describe in
Section~\ref{sec:finding_paths}. As stated in Appendix~\ref{sec:model_arch}, we added noise to both
the real and fake inputs of the discriminator in order to stabilize GAN~training on this dataset,
for which the pixels are binary-valued. Incorporating this modification into the training procedure
for InfoGAN may improve the score reported by~\citep{betaVAE}. Our approach achieves a score
competitive to that of~DC-IGN, which makes use of supervised information. We note that the encoder
trained to invert the generator does not always succeed in finding a latent variable that yields an
accurate reconstruction of the input. This may lead to a reduction in the disentanglement score, a
problem that does not affect the other approaches for which an accurate inference procedure is
available.

\begin{figure}[t]
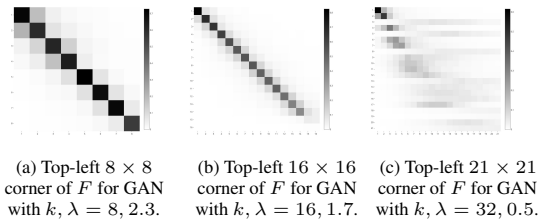

\centering
\newcommand{\image}[1]{\includegraphics[width=0.12\textwidth]{figures/heatmaps/#1}}

\begin{tabular}[t]{c@{\hspace{0.1in}}c@{\hspace{0.1in}}c}%
\captionsetup{width=.27\linewidth,justification=centering}%
\subfloat[][Top-left~${8 \times 8}$ corner of~$F$ for GAN with~${k, \lambda = 8, 2.3}$.]{%
	\image{{gan_nz_128_ac_8_aw_2.3}.jpg}
} &
\captionsetup{width=.31\linewidth,justification=centering}%
\subfloat[][Top-left~${16 \times 16}$ corner of~$F$ for GAN with~${k, \lambda = 16, 1.7}$.]{%
	\image{{gan_nz_128_ac_16_aw_1.7}.jpg}
} &
\captionsetup{width=.31\linewidth,justification=centering}%
\subfloat[][Top-left~${21 \times 21}$ corner of~$F$ for GAN with~${k, \lambda = 32, 0.5}$.]{%
	\image{{gan_nz_128_ac_32_aw_0.5}.jpg}
}
\end{tabular}
\caption{Comparison of the top-left corners of the matrix~$F$~(Equation~\ref{eq:heatmap_matrix}) for
alignment-regularized GANs with~${k \in \{8, 16, 32\}}$. All GANs were trained with latent variable
size~$128$. Details regarding the model architecture and training procedure are given in
Appendix~\ref{sec:model_arch}.\label{fig:gan_celeb_a_heatmaps}}
\end{figure}

\begin{figure}[ht]
\newcommand{\image}[1]{\includegraphics[width=0.20\textwidth]{figures/disentanglement/gan_ap_k_32/#1}}

\centering\tiny
\begin{tabular}{c l m{0.43\linewidth}}
\toprule
Coordinate & Description & Example \\
\midrule
1  & Background darkness        & \image{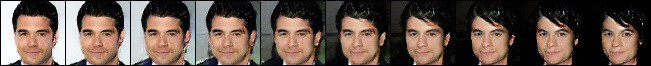}     \\
2  & Azimuth                    & \image{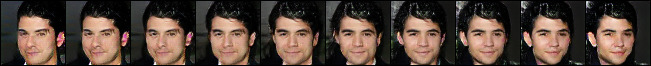}     \\
3  & Bangs                      & \image{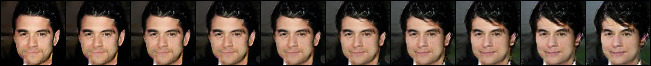}     \\
4  & Gender, bangs, and smiling & \image{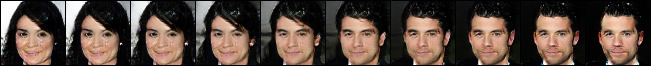}     \\
5  & Smiling                    & \image{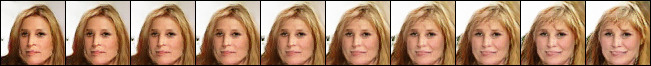}  \\
6  & Hair color                 & \image{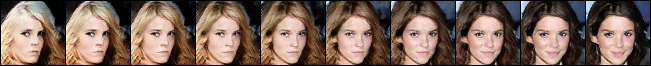}     \\
7  & Hair color and hair style  & \image{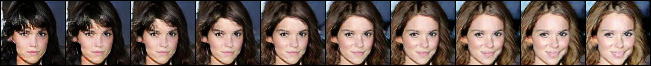}  \\
8  & Lighting color and bangs   & \image{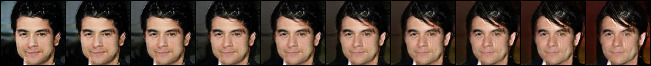}     \\
9  & Hair color and hair style  & \image{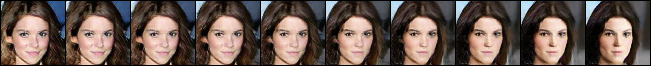}  \\
10 & Hair color and hair style  & \image{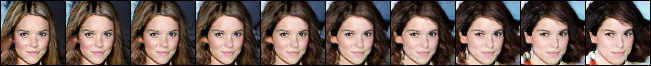}  \\
11 & Jawline                    & \image{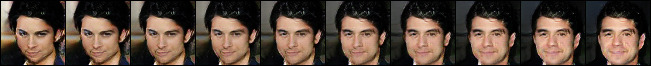}    \\
12 & Smiling and bangs          & \image{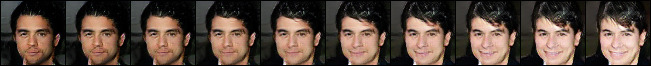}    \\
13 & Background color           & \image{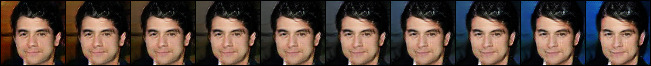}    \\
14 & Age and hairline           & \image{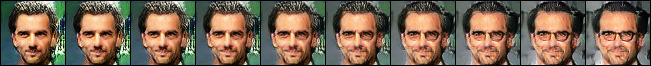}    \\
15 & Age                        & \image{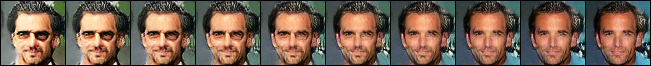} \\
17 & Location of hair partition & \image{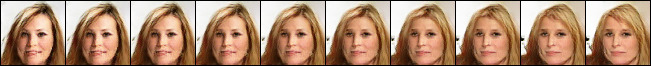}    \\
18 & Lighting and skin tone     & \image{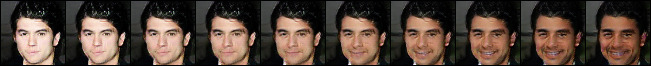}    \\
20 & Mouth open                 & \image{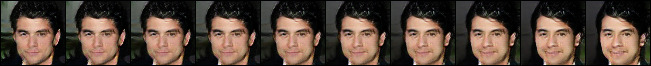}    \\
21 & Mouth open                 & \image{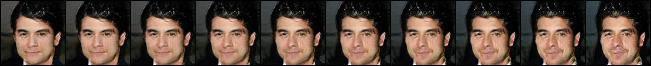}    \\
\bottomrule
\end{tabular}
\caption{Disentanglement results for alignment-regularized GAN~(${k, \lambda = 32, 0.5}$) with
latent variable size~$128$ on the CelebA dataset, at~${64 \times 64}$ resolution. Details regarding
the model architecture and training procedure are given in
Appendix~\ref{sec:model_arch}. Results with additional samples for each coordinate can be found at
this~URL: \url{https://drive.google.com/open?id=1E2TneLSAdyFYgN4GUzYKnHDYABo-G8RK}.%
\label{fig:gan_factors_32}}
\end{figure}

\begin{figure}[ht]
\newcommand{\image}[1]{\includegraphics[width=0.20\textwidth]{figures/disentanglement/vae_beta_8/#1}}

\centering\tiny
\begin{tabular}{c l m{0.43\linewidth}}
\toprule
Coordinate & Description & Example \\
\midrule
1  & Azimuth                         & \image{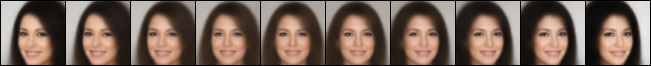} \\
2  & Background color                & \image{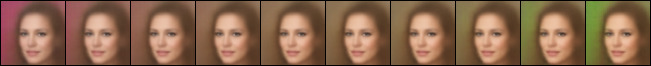} \\
3  & Location of hair partition      & \image{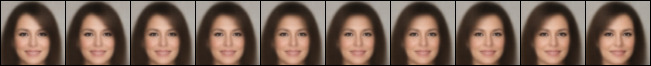} \\
4  & Skin color                      & \image{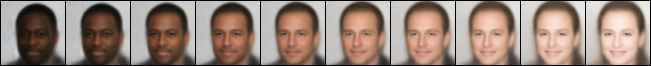} \\
5  & Hair direction                  & \image{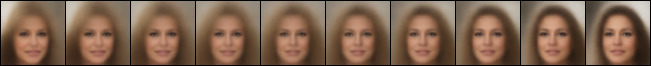} \\
8  & Background color                & \image{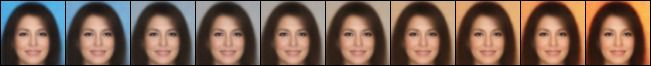} \\
12 & Hair length                     & \image{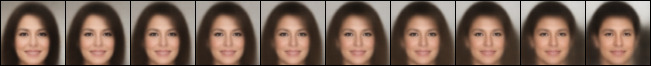} \\
14 & Jawline                         & \image{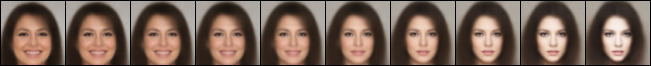} \\
19 & Skin tone                       & \image{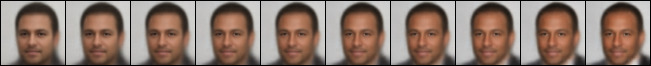} \\
20 & Hair style                      & \image{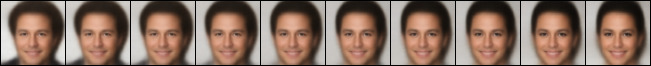} \\
22 & Sunglasses                      & \image{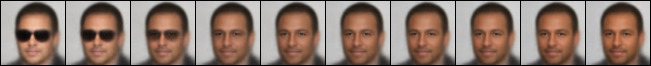} \\
23 & Gender                          & \image{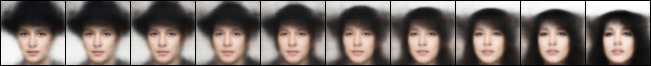} \\
25 & Background darkness             & \image{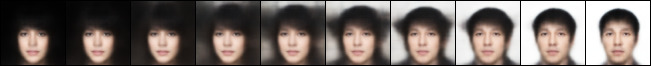} \\
26 & Hairline and skin tone          & \image{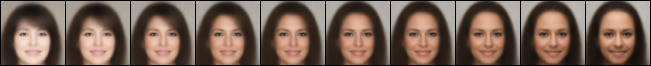} \\
28 & Lighting direction and rotation & \image{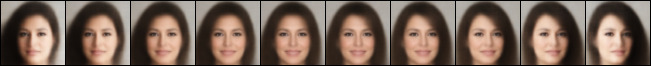} \\
32 & Hair size, color, and smiling   & \image{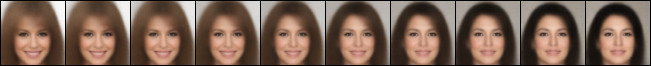} \\
\bottomrule
\end{tabular}
\caption{Disentanglement results for~$\beta$-VAE~(${\beta = 8}$) with latent variable size~$32$ on
the CelebA dataset, at~${64 \times 64}$ resolution. Details regarding the model architecture and
training procedure are given in Appendix~\ref{sec:model_arch}. Results with additional samples for
each coordinate can be found at this~URL:
\url{https://drive.google.com/open?id=1Zu5H_M0dOuE2NYeOJ9WZFSux4U5qM5Pq}.%
\label{fig:beta_vae_factors_8}}
\end{figure}

Next, we make a qualitative comparison between our approach and~$\beta$-VAE. We train a series
of~GANs on the CelebA dataset~\citep{CelebA}, with~$k \in \{{8, 16, 32}\}$ for
Algorithm~\ref{alg:alignment_reg} and varying values for the alignment regularizer weight~$\lambda$.
We also train a series of~$\beta$-VAEs with~${\beta \in \{1, 2, 4, 8, 16\}}$. The results for our
method with~${k, \lambda = 32, 0.5}$ are shown in Figure~\ref{fig:gan_factors_32}, and the results
for~$\beta$-VAE with~$\beta = 8$ in Figure~\ref{fig:beta_vae_factors_8}.
Figure~\ref{fig:gan_celeb_a_heatmaps} shows the top-left corners of the matrix~$F$ given by
Equation~\ref{eq:heatmap_matrix}, for three configurations of our approach with~$k \in \{{8, 16,
32\}}$. A penalty weight of~$\lambda = 0.5$ was not sufficient to result in a clear diagonal
structure in the top-left corner of the matrix~$F$ for the configuration with~$k = 32$.
Nonetheless, this configuration resulted in the largest number of disentangled factors, and the
results are shown in Figure~\ref{fig:gan_factors_32}. The best results for the configurations
with~${k \in \{8, 16\}}$ are shown in Appendix~\ref{sec:gan_disentanglement_extra}. We found that
the~$\beta$-VAE configuration with~${\beta = 8}$ resulted in the largest number of disentangled
factors, while still maintaining sufficiently low reconstruction error so as to keep the sample
quality acceptable. These results are shown in Figure~\ref{fig:beta_vae_factors_8}. In addition to
better sample quality, our approach is able to learn concepts such as different kinds of hair styles
(coordinates~9 and~10 in Figure~\ref{fig:gan_factors_32}) that are not modeled by the~$\beta$-VAE.

Even with relatively large values for the penalty weight, the heatmaps shown in
Figure~\ref{fig:gan_celeb_a_heatmaps} contain nonzero entries above and below the diagonal. This can
result in some degree of leakage of the attribute controlled by one coordinate into the next. To see
this in more detail, we examine the disentangled factors found by the configuration with~$k = 16$,
which are shown in Figure~\ref{fig:gan_factors_16}. Coordinates~${\{1, 2, 3\}}$ all involve change
in hair darkness; coordinates~${\{4, 5, 6\}}$ all involve change in gender; coordinates~$\{{6, 7, 8,
9}\}$ all involve smiling; and coordinates~$\{{12, 13\}}$ all involve change in the location of the
hair partition. This is a limitation of our current approach, and modifications to the
implementation of the alignment regularizer in Algorithm~\ref{alg:alignment_reg} may help to
mitigate this leakage. We plan to investigate such improvements in future work.

\section{Conclusion}

Our work approaches the problem of learning disentangled representations from the perspective of
eigenvector alignment. We develop a novel regularizer which, when incorporated into the GAN
objective, induces disentangled representations of quality comparable to those obtained by VAE-based
approaches~(\citet{betaVAE}, \citet{FactorVAE}, \citet{TCVAE}, \citet{HFVAE}), without the need to
introduce auxiliary models into the training procedure~\citep{InfoGAN}. This approach is also not
specific to the GAN framework~\citep{GAN}, and could potentially be applied to autoregressive
models, such as those used to generate text and audio. We believe this is an important direction for
future investigation. So far, two different perspectives for viewing disentanglement have been
proposed: maximizing mutual information and eigenvector alignment. An investigation into the
relationship between the two could further our understanding of what disentanglement is and why it
occurs.

\section*{Acknowledgements}

We would like to thank for Ryan-Chris Moreno Vasquez and Emily Denton for early discussions that led
us to think carefully about the paths determined by the leading eigenvectors of~$M_z(z_0)$. We are
also grateful for the suggestions provided by Mikael Henaff, Alec Radford, Yura Burda, and Harri
Edwards, which improved the quality of our presentation.

\clearpage 
\bibliography{main}
\bibliographystyle{icml2019}

\clearpage\onecolumn\appendix
\section{Forward- and reverse-mode automatic differentiation}\label{sec:autodiff}

The entries of~$\jg$ are not explicitly stored in memory: it is an implicit-defined matrix.
As such, information about~$\jg$ must be accessed via automatic differentiation~(AD), of which there
are two kinds: forward-mode and reverse-mode. We briefly describe them here, and refer the reader
to~\citep{ADSurvey} for a more comprehensive survey.

Suppose that we are given a differentiable function~${f : \reals^m \to \reals^n}$ corresponding to a
feedforward neural network with~$L$ layers, and let~${G \coloneqq (V, E)}$ be its representation as
a computation graph. For each~${k \in [1, L]}$, let~$R_k$ be the set of nodes corresponding to
the~$k$th layer, and let~${r_k : \reals^{\alpha_{k - 1}} \to \reals^{\alpha_k}}$ be the function
computed by the nodes in~$R_k$, with~${\alpha_0 \coloneqq m}$ and ${\alpha_L \coloneqq n}$. Then~${f
= r_L \circ r_{L - 1} \circ \cdots \circ r_2 \circ r_1}$. Finally, let~${\bar{x} \in \reals^m}$
denote the variable for the input to~$f$, and~${\bar{b}_{k - 1} \in \reals^{\alpha_{k - 1}}}$ the
variable for the input to~$r_k$, with~${\bar{b}_0 \coloneqq \bar{x}}$. Given a vector~${v \in
\reals^m}$, forward-mode~AD computes~$Jv$. It works according to the recurrence
\begin{align*}
	v_k \coloneqq \left( D_{\bar{x}} r_k \big|_{\bar{x} = x} \right) v
	&= \left( D_{\bar{b}_{k - 1}} r_k \big|_{\bar{b}_{k - 1} = b_{k - 1}} \right)
	   \left( D_{\bar{x}}\; r_{k - 1} \circ \ldots \circ r_1 \big|_{\bar{x} = x} \right) v \\
	&= \left( D_{\bar{b}_{k - 1}} r_k \big|_{\bar{b}_{k - 1} = b_{k - 1}} \right) v_{k - 1},
\end{align*}
where~$k \in [1, L]$ and~$v_0 \coloneqq v$. After step~$k$, we will have obtained the product of the
Jacobian of~$r_k$ with respect to~$x$, and~$v$; after step~$L$, we will have obtained the desired
product~$Jv$.

Forward-mode~AD computes~$Jv$ far more efficiently than the approach of first evaluating~$J$, and
subsequently multiplying by~$v$. Suppose for simplicity that~${\alpha_k = n}$ for all~${k \in [0,
L]}$, so that~${J \in \reals^{n \times n}}$. To compute~$J$ independently, we would use the chain
rule, which gives
\[
	D_{\bar{x}} f \big|_{\bar{x} = x} =
		\left( D_{\bar{b}_{L - 1}} r_L \big|_{\bar{b}_{L - 1} = b_{L - 1}} \right)
		\left( D_{\bar{b}_{L - 2}} r_{L - 1} \big|_{\bar{b}_{L - 2} = b_{p - 2}} \right)
		\cdots \\
		\left( D_{\bar{b}_0} r_1 \big|_{\bar{b}_0 = x} \right).
\]
Each layer~$r_k$, for~${k \in [1, L]}$, must compute its~$n \times n$ Jacobian, and each layer after
the first must multiply its Jacobian with the~$n \times n$ Jacobian of the preceding layer with
respect to~$x$. This process is described by the recurrence
\[
	D_{\bar{x}} r_k \big|_{\bar{x} = x} =
		\left( D_{\bar{b}_{k - 1}} r_k \big|_{\bar{b}_{k - 1} = b_{k - 1}} \right)
		\left( D_{\bar{x}} r_{k - 1} \big|_{\bar{x} = x} \right).
\]
Assuming that~$\Theta(1)$ operations are required to compute each element of the Jacobian, the total
number of operations required is proportional to
\[
	N \coloneqq n^2 + (L - 1) (n^2 + n^3) \in \Theta(n^3),
\]
despite the fact that~$J$ has only~$n^2$ elements. On the other hand, the~$k$th step of the
forward-mode recurrence requires~$\Theta(n)$ operations, so computing~$Jv$ only
requires~$\Theta(Ln)$ operations. By running forward-mode with~${v = e_i}$ for each~$i \in [1, n]$,
we can form~$J$ column-by-column in only~$\Theta(Ln^2)$ operations. Henceforth, we will measure
operation count in terms of invocations to the~AD engine, rather than by counting elementary
operations.

Of the two kinds of~AD, it is reverse-mode that finds the most use in machine learning. Given a
vector~${w \in \reals^n}$, it computes~$J^t w$. Most applications involve minimizing a scalar-valued
loss function with respect to a vector of parameters, which corresponds to the case where~$n = 1$.
This special case is otherwise known as backpropagation. Unlike forward-mode, which begins at the
first layer and ends at the last, reverse-mode begins at the last layer and ends at the first. It
works according to the recurrence
\begin{align*}
	w_k^t \coloneqq w^t \left( D_{\bar{b}_{k - 1}} r_k \big|_{\bar{b}_{k - 1} = b_{k - 1}} \right)
	&= w^t \left( D_{\bar{b}_k}\; r_L \circ \cdots \circ r_{k + 1} \big|_{\bar{b}_k = b_k} \right)
	       \left( D_{\bar{b}_{k - 1}} r_k \big|_{\bar{b}_{k - 1} = b_{k - 1}} \right) \\
	&= w_{k + 1}^t \left( D_{\bar{b}_{k - 1}} r_k \big|_{\bar{b}_{k - 1} = b_{k - 1}} \right), \\
\end{align*}
where~$k$ ranges from~$L$ to~$1$, and~$w_L \coloneqq w$. After~$L$ steps, we will have obtained the
desired product~$J^t w$. Machine learning frameworks typically expose the full interface to the
reverse-mode~AD engine, rather than specializing to the case of backpropagation. E.g., in
TensorFlow~\citep{TF}, one can change the value of~$w$ for \lstinline{tf.gradients} from a vector of
ones to a custom value specified by the \lstinline{grad_ys} parameter.

\begin{figure}[t]
\begin{lstlisting}[style=pythonStyle, caption={TensorFlow implementation of Jacobian-vector
operations.}, captionpos=b, abovecaptionskip=\medskipamount, label={lst:jacobian_ops}]
import tensorflow as tf

def forward_gradients(ys, xs, d_xs):
	v = tf.placeholder_with_default(tf.ones_like(ys), shape=ys.get_shape())
	g = tf.gradients(ys, xs, grad_ys=v)
	return tf.gradients(g, v, grad_ys=d_xs)

def j_v(ys, xs, vs):
	return forward_gradients(ys, xs, vs)

def jt_v(ys, xs, vs):
	return tf.gradients(ys, xs, vs)

def jt_j_v(ys, xs, vs):
	jv = j_v(ys, xs, vs)
	return tf.gradients(ys, xs, jv)

def j_jt_v(ys, xs, vs):
	jt_v_ = jt_v(ys, xs, vs)
	return forward_gradients(ys, xs, jt_v_)
\end{lstlisting}
\end{figure}

If desired, we can compute the entire Jacobian using either type of~AD. By running forward-mode
with~$v = e_i$ for each~$i \in [1, m]$, we can form~$J$ column-by-column, using~$m$ total
invocations to~AD. Similarly, by running reverse-mode with~$w = e_i$ for each~$i \in [1, n]$, we can
form~$J$ row-by-row using~$n$ total invocations to~AD. If~$n > m$, the former is typically faster;
otherwise, the latter is preferable. We note that while~AD offers a relatively efficient approach
for evaluating~$J$, doing so at each iteration of optimization becomes impractical.

Many popular machine learning frameworks (e.g., TensorFlow) do not implement forward-mode natively,
since it is seldom used for machine learning. Surprisingly, this is not a limitation:
\emph{reverse-mode can be used to implement forward-mode.} Given a differentiable function~${f :
\reals^m \to \reals^n}$, we can compute~$w^t J$ for a given vector~$w \in \reals^n$, using
reverse-mode AD. Treating the input to~$f$ as a constant, we can regard~$w^t J$ as a function~${g :
\reals^n \to \reals^m}$, $w \mapsto w^t J$. The derivative of~$g$ with respect to~$w$ is given
by~$J^t$, and so another application of reverse-mode AD allows us to compute~${v^t J^t = (Jv)^t}$.
Hence, reverse-mode AD can be used to implement forward-mode AD. This trick was first described
by~\citep{jvp_from_vjp}. We provide a TensorFlow implementation of the procedures to compute~$Jv,
J^t v, J^t J v$, and~$J J^t v$ in Listing~\ref{lst:jacobian_ops}.

\section{Eigenvector Alignment for~$\beta$-VAEs}
\label{sec:beta_vae_alignment}

\begin{figure}[t]
\centering
\newcommand{\halfwidthimage}[1]{\includegraphics[width=0.30\textwidth]{figures/heatmaps/#1}}
\subfloat[][]{\halfwidthimage{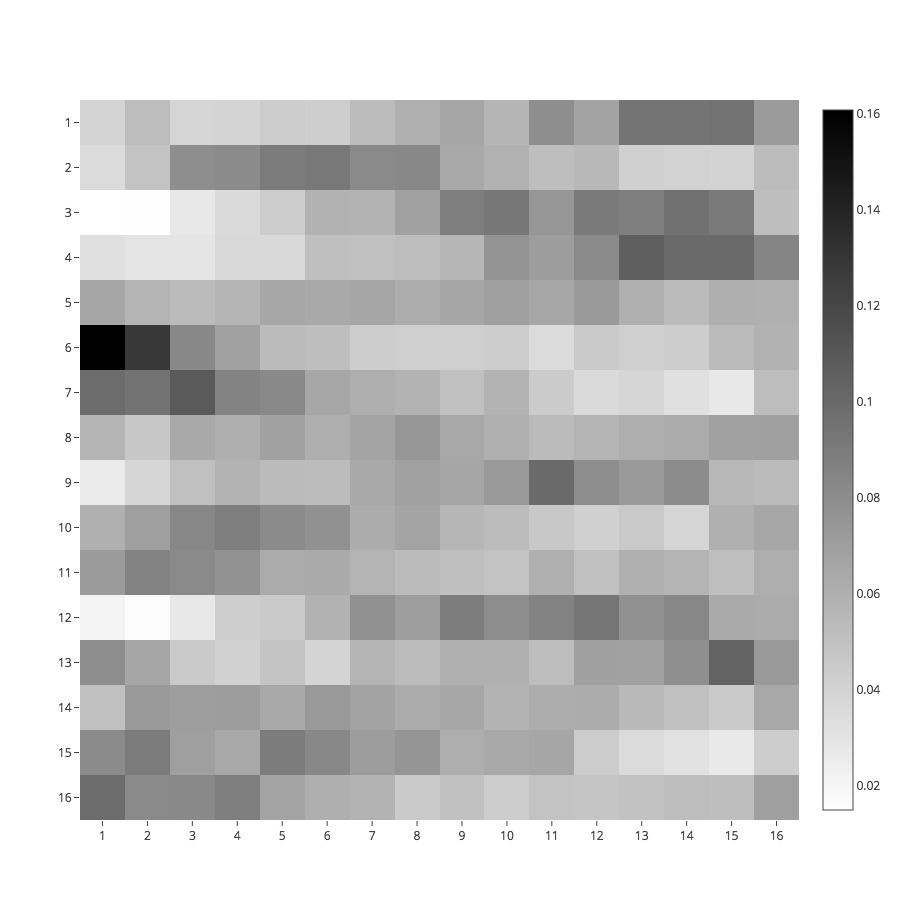}}
\hspace{0.2in}
\subfloat[][]{\halfwidthimage{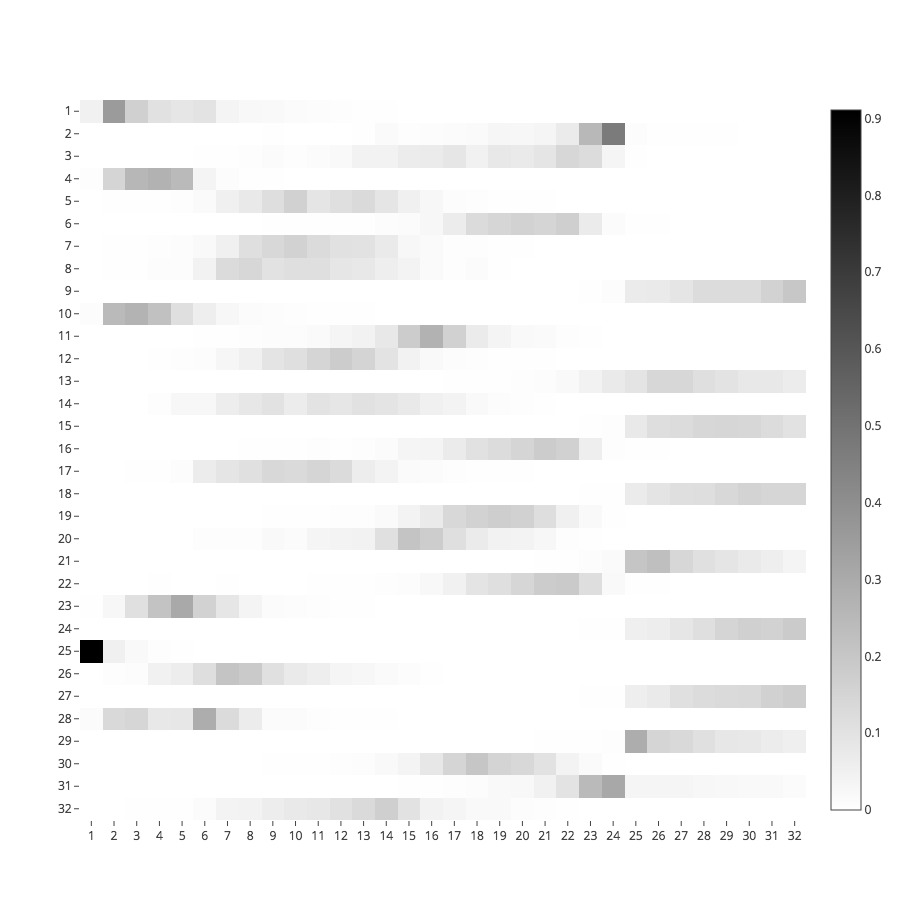}}
\caption{Comparison between the average squared eigenvector matrices~$F$
(Equation~\ref{eq:heatmap_matrix}) for a~GAN generator with latent variable size~$16$~(left) and the
decoder from a~$\beta$-VAE~(${\beta = 8}$) with latent variable size~$32$~(right), both trained on
the CelebA dataset. Complete architecture and training details are provided in
Appendix~\ref{sec:model_arch}. The matrix~$F$ for the~VAE decoder contains several columns that are
close to one-hot vectors. By contrast, the same matrix for the~GAN generator exhibits little
structure, despite the fact that it is trained with a small latent variable
size.\label{fig:generator_heatmaps}}
\end{figure}

\begin{figure}[t]
\centering
\newcommand{\fullwidthimage}[1]{\includegraphics[width=0.7\textwidth]{figures/heatmaps/#1}}
\fullwidthimage{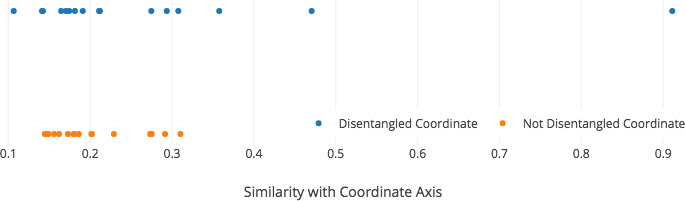}
\caption{Investigation of whether the matrix~$F$ can be used to determine which directions in latent
space correspond to disentangled factors for a~$\beta$-VAE. For each~$i \in [1, 32]$, we compute the
maximum value along the~$i$th row of matrix~$F$ shown in Figure~\ref{fig:generator_heatmaps}(b).
Then, we plot a point indicating whether or not disentanglement occurs along this direction, as
determined by visual inspection. The directions that do result in disentanglement are shown in
Figure~\ref{fig:beta_vae_factors_8}.\label{fig:beta_vae_heatmap_disentanglement}}
\end{figure}

The results from Section~\ref{sec:finding_paths} suggest that alignment of the eigenvectors
of~$M_z(z_0)$ with the coordinate axes might be sufficient to induce disentanglement in the latent
representation of a generative model. The~$\beta$-VAE is known to learn such a representation
when~$\beta$ is increased, so that the KL-divergence between the approximate posterior and the prior
of the decoder is made sufficiently small. Suppose that a~$\beta$-VAE exhibits disentanglement along
the~$j$th coordinate direction in latent space. Then paths of the form~${\gamma : t \mapsto z_0 + t
e_j}$ will produce changes to~$G(z_0)$ along isolated factors of variation. If such a path coincides
with the trajectory found by Algorithm~\ref{alg:trace_eigenpath} for the~$k$th leading eigenvector
throughout latent space, then this eigenvector must be aligned with the~$j$th coordinate axis.
Hence, we would expect the~$k$th column of the matrix~$F$ given by Equation~\ref{eq:heatmap_matrix}
to be a one-hot vector close to~$e_j$. Figure~\ref{fig:generator_heatmaps} shows that, in fact,
several columns of~$F$ have high similarity to coordinate directions.

Next, we investigate whether a column of~$F$ having high similarity with a coordinate
direction~$e_j$ implies that disentanglement actually occurs along~${\gamma : t \mapsto z_0 + t
e_j}$. Figure~\ref{fig:beta_vae_heatmap_disentanglement} shows that with the exception of three
coordinates having similarity greater than~0.35, this turns out not to be the case. In other words,
several of the eigenvectors naturally align with coordinate directions during training, but
disentanglement does not reliably occur along all of them. More work needs to be done in order to
better understand the relationship between eigenvector alignment and disentangled representations.

\section{Generator Architectures and Training Procedure} \label{sec:model_arch}

\begin{table}[t]
\centering
\footnotesize
\begin{tabular}{cccc}
\toprule
Figure & DCGAN Base Feature Map Count & Latent Variable Size & Notes \\
\midrule
\ref{fig:latent_perturbations}(c)          & 64  & 32  & 1 \\
\ref{fig:latent_perturbations}(e)          & 64  & 64  & 1 \\
\ref{fig:latent_perturbations}(g)          & 64  & 128 & 1 \\
\ref{fig:latent_perturbations}(d),%
\ref{fig:alignment_heatmaps}, %
\ref{fig:alignment_reg_perf}, %
\ref{fig:gan_celeb_a_heatmaps}(a), %
\ref{fig:gan_celeb_a_heatmaps}(b), %
\ref{fig:gan_factors_32}                   & 64  & 128 & 1 \\
\ref{fig:eigenpaths_celeb_a}, %
\ref{fig:gan_factors_8}, %
\ref{fig:gan_factors_16}                   & 128 & 128 & 1 \\
\ref{fig:latent_perturbations}(f)          & 64  & 256 & 1 \\
\ref{fig:latent_perturbations}(h)          & 64  & 512 & 1 \\
\ref{fig:eigenpaths_dsprites}              & 64  & 3   & 1, 3 \\
\ref{table:disentanglement_score_results}  & 64  & 10  & 1, 3 \\
\ref{fig:generator_heatmaps}(a)            & 64  & 16  & 1 \\
\ref{fig:generator_heatmaps}(b), %
\ref{fig:beta_vae_heatmap_disentanglement} & 64  & 32  & 2 \\
\bottomrule
\end{tabular}
\caption{GAN and~$\beta$-VAE architectures used for all figures and tables. Note~1: weight
normalization~\citep{WeightNorm} was applied both to the generator and the discriminator, with the
scale~$g$ fixed to one for the discriminator. Note~2: spectral weight
normalization~\citep{SpectralWeightNorm} was applied both to the encoder and the decoder, with
learned scales for both. Note~3: gaussian noise with standard deviation~$0.6$ was added to both real
and fake inputs to the discriminator.\label{table:model_arch}}
\end{table}

All models in this work are based on the~DCGAN~\citep{DCGAN} architecture.
Table~\ref{table:model_arch} describes the architectures of the generator and discriminator~(in the
case of~GANs) and the encoder and decoder~(in the case of~VAEs) associated with each figure and
table. All models use the translated PReLU activation function~\citep{TPReLU}; the ReLU leaks are
learned, and clipped to the interval~$[0, 1]$ after each parameter update. The only data
preprocessing we applied was to scale the pixel values to the interval~$[0, 1]$. The GANs were
trained using the original, non-saturating GAN loss described in~\cite{GAN} with multivariate normal
prior. Both the generator and the discriminator were trained using RMSProp~\citep{RMSProp} with step
size~$10^{-4}$, decay factor~$0.9$, and~${\epsilon = 10^{-6}}$. Each parameter update was made using
a batch size of~$32$, and the models were trained for a total of~\num{750000} updates. The VAEs were
trained using a gaussian likelihood model for the decoder, and the log-diagonal covariance
parameterization for the encoder. The fixed, per-pixel standard deviation of the decoder was chosen
such that, disregarding constant terms, the log-likelihood corresponds to the reconstruction error,
normalized by the product of the number of channels and the number of pixels. To optimize the
evidence lower bound, we used Adam~\citep{Adam} with~${\beta_1, \beta_2, \epsilon = 0.5, 0.99,
10^{-8}}$. Each parameter update was made using a batch size of~$32$, and the models were trained
for a total of~\num{1500000} updates. The KL-divergence weight~$\beta$ was increased linearly
from~$0$ to the final value over the first~\num{1000000} updates. We applied the function~$x \mapsto
2 (\tanh(x) + 1 / 2)$ to the outputs of both the~GAN generators and the~VAE decoders; this ensures
that the output pixel values are in the interval~$[0, 1]$. To predict the mean and log-diagonal
covariance with the VAE~encoders, we apply the translated ReLU activation function~\citep{TPReLU} to
the final convolutional features, followed by two fully-connected layers, one for each statistic.

\section{Implementation Details for Disentanglement Metric Score}
\label{sec:disentanglement_metric_score}

\begin{algorithm}[t]\tiny
\caption{Procedure to generate a batch for disentanglement metric
classifier.\label{alg:classifier_batch}}
\begin{algorithmic}
\Require{$n_\text{inst} \geq 1$ is the number of aligned pairs to use, in order to create each
instance in the batch.}
\Require{$n_\text{batch} \geq 1$ is the batch size.}
\Require{$\enc : \reals^n \to \reals^m$ is the encoder functioning as the inverse of the
generator~$G$.}
\State

\State{$c \gets (3, 6, 40, 30, 30)$} \Comment{Ranges for the five attributes that determine each
shape}
\State

\Procedure{SampleShape}{$i, v$} \Comment{Sample shape with attribute~$i$ fixed to value~$v$}
	\For{$j \in [1, 5]$}
		\State{$u_j \gets \Call{RandomUniform}{1, c_j}$}
	\EndFor
	\State{$u \gets (u_1, u_2, u_3, u_4, u_5)$}
	\State{$u_i \gets v$}
	\State \Return $\Call{MakeShape}{u}$
\EndProcedure
\State

\Procedure{MakeInstance}{$n_\text{inst}, \enc$}
	\State{$i \gets \Call{RandomUniform}{2, 5}$} \Comment{Sample index of attribute to fix}
	\State{$v \gets \Call{RandomUniform}{1, c_i}$} \Comment{Sample value for fixed attribute}
	\State{$z \gets 0 \in \reals^m$}
	\For{$k \in [1, n_\text{inst}]$}
		\State{$z_0 \gets \enc(\Call{SampleShape}{i, v})$}
		\State{$z_1 \gets \enc(\Call{SampleShape}{i, v})$}
		\State{$z \gets z + |z_0 - z_1|$}
	\EndFor
	\State \Return $z / n_\text{inst}, i$
\EndProcedure
\State

\Procedure{MakeBatch}{$n_\text{inst}, n_\text{batch}, \enc$}
	\State{$\text{inputs} \gets \varnothing$}
	\State{$\text{targets} \gets \varnothing$}
	\For{$k \in [1, n_\text{batch}]$}
		\State{$x, y \gets \Call{MakeInstance}{n_\text{inst}, \enc}$}
		\State{$\text{inputs} \gets \text{inputs} \cup \{x\}$}
		\State{$\text{targets} \gets \text{targets} \cup \{y\}$}
	\EndFor
	\State \Return $\text{inputs}, \text{targets}$
\EndProcedure

\end{algorithmic}
\end{algorithm}

Algorithm~\ref{alg:classifier_batch} describes the procedure used to generate the batches to train
and evaluate the classifier for the disentanglement metric~\citep{betaVAE}. We use~${n_\text{inst},
n_\text{batch} = 64, 32}$, and update the classifier using~SGD with nesterov accelerated
gradient~(step size~$10^{-2}$, momentum~$0.99$). We train the classifier using a total
of~\num{10000} parameter updates, and evaluate its performance using~\num{5000} instances, as
reported by~\citep{betaVAE}. To invert the pretrained GAN generator, we train a~VAE decoder with
twice the base feature map count for the generator, using~\num{30000} parameter updates. The details
for this training procedure are identical to those for regular VAE training described in
Appendix~\ref{sec:model_arch}, except that the generator is not updated, and the KL-divergence
weight~$\beta$ is set to zero. In other words, we use a standard autoencoding loss with a fixed
decoder.

\clearpage
\section{Additional GAN Disentanglement Results} \label{sec:gan_disentanglement_extra}

\begin{figure}[ht]
\newcommand{\image}[1]{\includegraphics[width=0.40\textwidth]{figures/disentanglement/gan_ap_k_8/#1}}
\centering\footnotesize
\begin{tabular}{c l m{0.40\textwidth}}
\toprule
Coordinate & Description & Example \\
\midrule
1 & Background darkness and hair color & \image{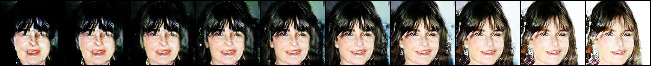} \\
2 & Azimuth, lighting, and hair color  & \image{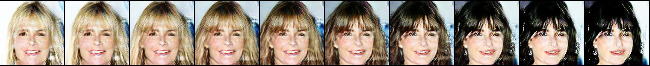} \\
3 & Hairline and hair color            & \image{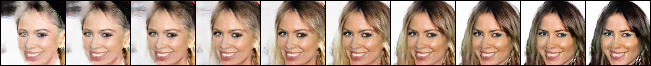} \\
4 & Azimuth                            & \image{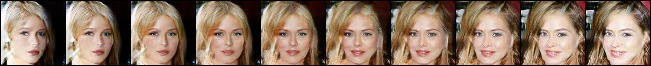} \\
5 & Shadow                             & \image{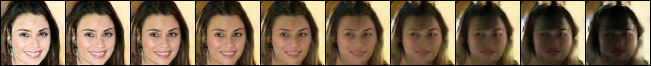} \\
6 & Smiling, age, skin tone, gender    & \image{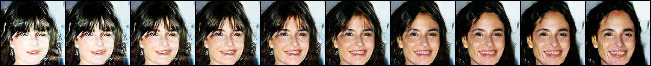} \\
7 & Smiling, age, jawline              & \image{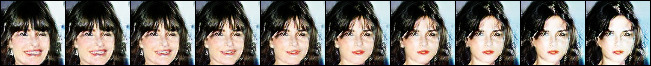} \\
8 & Jawline and hairstyle              & \image{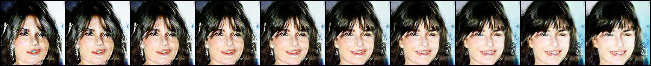} \\
\bottomrule
\end{tabular}
\captionsetup{font=footnotesize}
\caption{Disentanglement results for alignment-regularized GAN~(${k, \lambda = 8,
2.3}$). Details regarding the model architecture and training procedure are given in
Appendix~\ref{sec:model_arch}. Results with additional samples for each coordinate can be found at
this~URL: \url{https://drive.google.com/open?id=1Wnd5jIxopFsBRlylMUN2HfWhICXGiU3u}.%
\label{fig:gan_factors_8}}
\end{figure}

\begin{figure}[ht]
\newcommand{\image}[1]{\includegraphics[width=0.40\textwidth]{figures/disentanglement/gan_ap_k_16/#1}}
\centering\footnotesize
\begin{tabular}{c l m{0.40\textwidth}}
\toprule
Coordinate & Description & Example \\
\midrule
1  & Background and hair darkness                   & \image{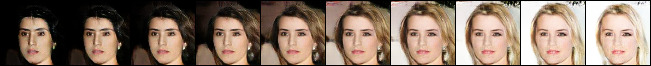}     \\
2  & Azimuth and hair darkness                      & \image{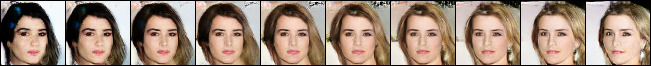}     \\
3  & Hair length, hair darkness, and lighting       & \image{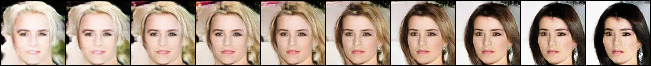}     \\
4  & Smiling and gender                             & \image{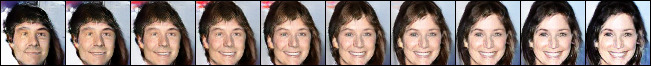}     \\
5  & Hair length, hair darkness, and gender         & \image{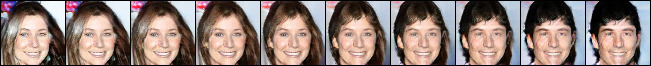}     \\
6  & Smiling, bangs, and gender                     & \image{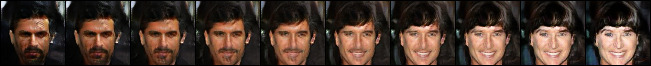}     \\
7  & Smiling and hairstyle                          & \image{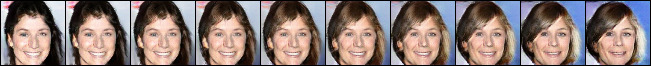}     \\
8  & Smiling, jawline, and glaring expression       & \image{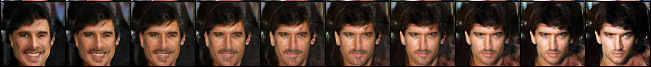}     \\
9  & Smiling, bangs, and mouth open                 & \image{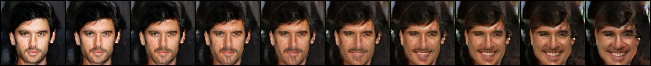}     \\
10 & Hairline                                       & \image{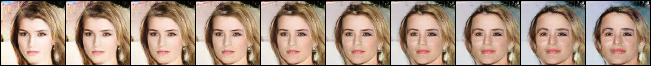}     \\
11 & Raised eyebrows and skin tone                  & \image{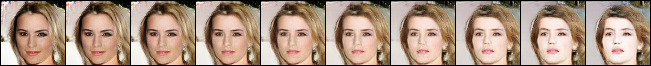}    \\
12 & Raised eyebrows and location of hair partition & \image{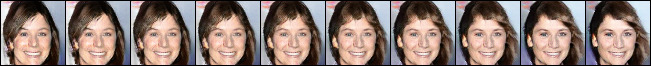} \\
13 & Raised eyebrows and location of hair partition & \image{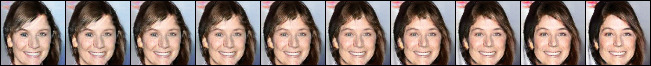}    \\
14 & Lighting color                                 & \image{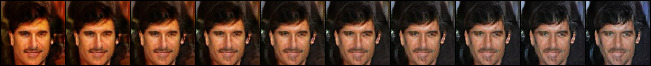}    \\
\bottomrule
\end{tabular}
\captionsetup{font=footnotesize}
\caption{Disentanglement results for alignment-regularized GAN~(${k, \lambda = 16,
1.7}$). Details regarding the model architecture and training procedure are given in
Appendix~\ref{sec:model_arch}. Results with additional samples for each corodinate can be found at
this~URL: \url{https://drive.google.com/open?id=1BM-P-hMF7sV_0smNFD_iTAKpq6FeNsCo}.%
\label{fig:gan_factors_16}}
\end{figure}

\clearpage
\section{Supplementary Figures}\label{sec:supplementary_figures}

\begin{figure}[t]
\newcommand{\fullwidthimage}[1]{\includegraphics[width=\textwidth]{figures/local_factors/#1}}
\newcommand{\halfwidthimage}[1]{\includegraphics[width=0.47\textwidth]{figures/local_factors/#1}}

\centering
\subfloat[][Log spectra for CelebA models with~$n_f = 64$ and~$m$ varied.]{%
	\fullwidthimage{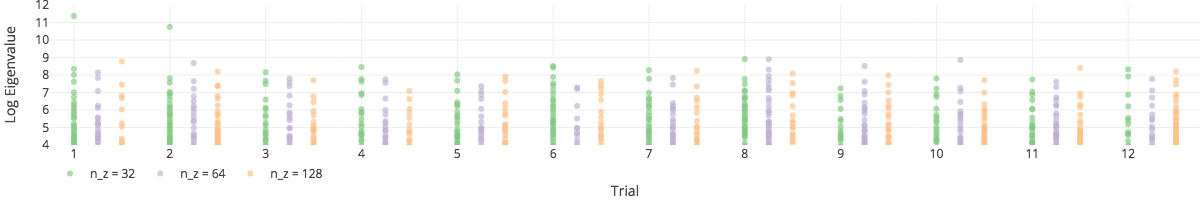}
} \\
\subfloat[][Log spectra for LSUN Bedroom models with~$n_f = 256$ and~$m$ varied.]{%
	\fullwidthimage{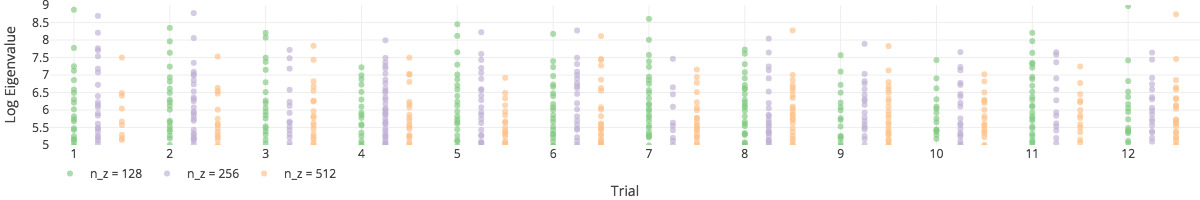}
} \\
\begin{tabular}[t]{c@{}c}%
\captionsetup{width=.49\linewidth,justification=centering}%
\subfloat[][CelebA ($m = 32, n_f = 64, \epsilon = 0.40$), embeddings~$z_{10}, z_{11}$.]{%
	{\def\arraystretch{0}%
	\begin{tabular}[b]{c}%
	\halfwidthimage{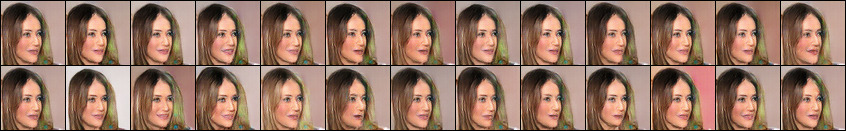} \\
	\halfwidthimage{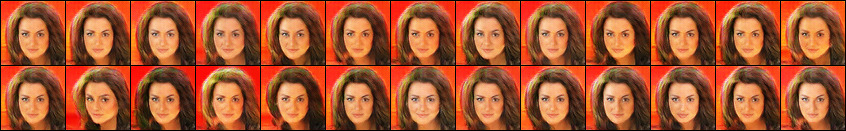}
	\end{tabular}}%
} &
\captionsetup{width=.49\linewidth,justification=centering}%
\subfloat[][LSUN Bedroom ($m = 128, n_f = 64, \epsilon = 0.80$), embeddings~$z_{5}, z_{12}$.]{%
	{\def\arraystretch{0}%
	\begin{tabular}[b]{c}%
	\halfwidthimage{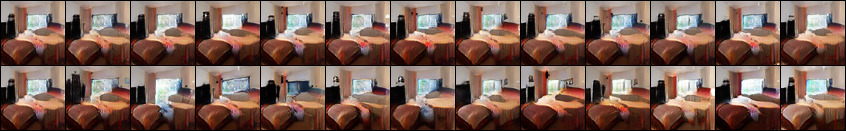} \\
	\halfwidthimage{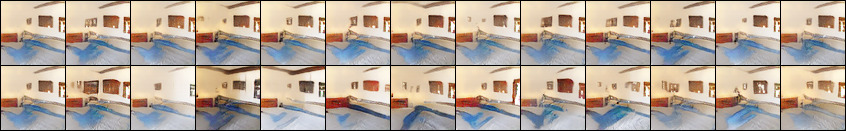}
	\end{tabular}}%
} \\
\captionsetup{width=.49\linewidth,justification=centering}%
\subfloat[][CelebA ($m = 64, n_f = 64, \epsilon = 0.65$), embeddings~$z_6, z_{12}$.]{%
	{\def\arraystretch{0}%
	\begin{tabular}[b]{c}%
	\halfwidthimage{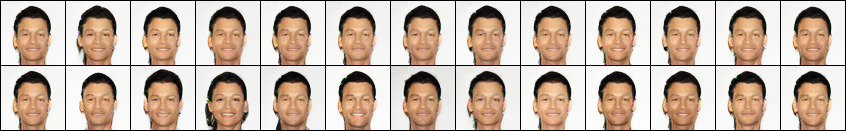} \\
	\halfwidthimage{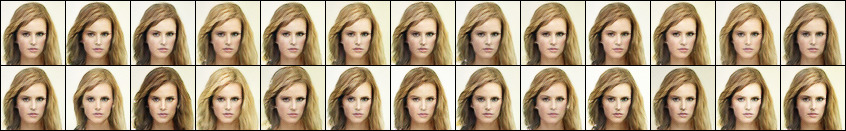}
	\end{tabular}}%
} &
\captionsetup{width=.49\linewidth,justification=centering}%
\subfloat[][LSUN Bedroom ($m = 256, n_f = 64, \epsilon = 0.80$), embeddings~$z_6, z_8$.]{%
	{\def\arraystretch{0}%
	\begin{tabular}[b]{c}%
	\halfwidthimage{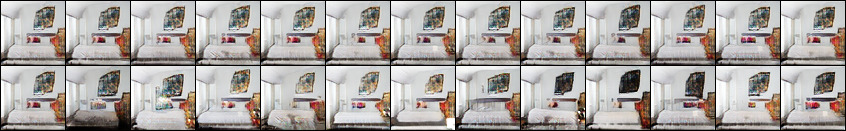} \\
	\halfwidthimage{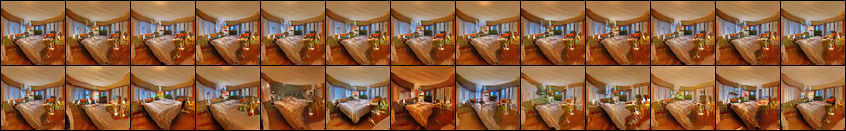}
	\end{tabular}}%
} \\
\captionsetup{width=.49\linewidth,justification=centering}%
\subfloat[][CelebA ($m = 128, n_f = 64, \epsilon = 0.80$), embeddings~$z_7, z_9$.]{%
	{\def\arraystretch{0}%
	\begin{tabular}[b]{c}%
	\halfwidthimage{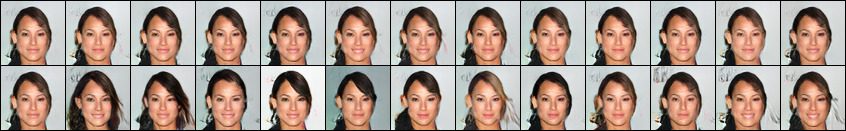} \\
	\halfwidthimage{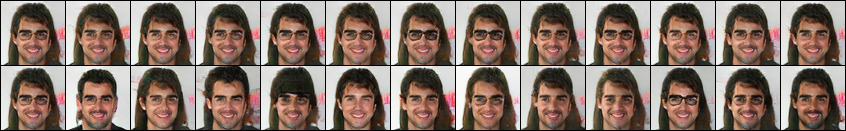}
	\end{tabular}}%
} &
\captionsetup{width=.49\linewidth,justification=centering}%
\subfloat[][LSUN Bedroom ($m = 512, n_f = 64, \epsilon = 0.80$), embeddings~$z_8, z_{10}$.]{%
	{\def\arraystretch{0}%
	\begin{tabular}[b]{c}%
	\halfwidthimage{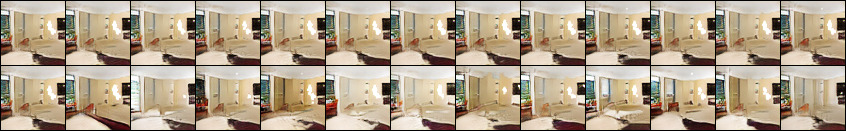} \\
	\halfwidthimage{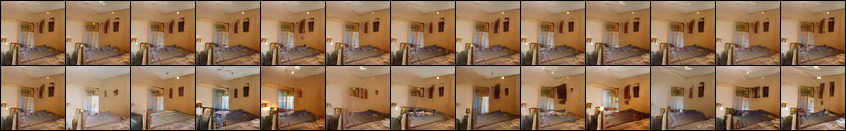}
	\end{tabular}}%
}%
\end{tabular}
\caption{Effect of perturbing a latent variable~$z$ along the leading eigenvectors of~$M_z(z)$.
Subfigures~(a) and~(b) show the top eigenvalues of $M_z(z_i)$, where~$\{z_1, \ldots, z_{12}\}$ are
fixed embeddings; small eigenvalues are omitted. Subfigures~(c)--(d) compare the effects of
perturbations along random directions to perturbations along leading eigenvectors. Each subfigure
consists of two stacked two-row grids. The leftmost images of each grid are both identical and equal
to~$G(z_i)$, for some~$i \in [1, 12]$. The first row shows the effect of perturbing~$z_i$ along
13~directions sampled uniformly from the sphere of radius~$\epsilon$, while the second row shows the
effect of perturbing~$z_i$ along the first 13~leading eigenvectors of~$M_z(z_i)$, by the same
distance~$\epsilon$. Details regarding the model architectures are given in
Appendix~\ref{sec:model_arch}.\label{fig:latent_perturbations}}
\end{figure}
\clearpage

\end{document}